\documentclass{article}

\usepackage{PRIMEarxiv}

\usepackage{algorithm}
\usepackage{algpseudocode}
\usepackage{amsfonts}
\usepackage{booktabs}       % professional-quality tables
\usepackage[utf8]{inputenc} % allow utf-8 input
\usepackage[T1]{fontenc}    % use 8-bit T1 fonts
\usepackage{hyperref}       % hyperlinks
\usepackage[nolist]{acronym}
\usepackage{url}            % simple URL typesetting
\usepackage{amsfonts}       % blackboard math symbols
\usepackage{nicefrac}       % compact symbols for 1/2, etc.
\usepackage{microtype}      % microtypography
\usepackage{lipsum}
\usepackage{fancyhdr}       % header
\usepackage{float}
\usepackage{multirow}
\usepackage{graphicx}       % graphics
\usepackage{subcaption}

\usepackage{bm}
\graphicspath{{media/}}     % organize your images and other figures under media/ folder

\title{Impact of HPO on AutoML Forecasting Ensembles
\thanks{\textit{\underline{Citation}}: 
\textbf{D. Hoffmann. Impact of HPO on AutoML Forecasting Ensembles}} 
}

\author{
  David B. Hoffmann \\
  Amazon Research \\
  Germany \\
  \texttt{adavidho@amazon.com} \\
}

\begin{document}
\maketitle
% Loads acronyms but doesn't print them
\begin{acronym}[XXXXXXX]
    \acro{ARIMA}{Autoregressive Integrated Moving Average}
    \acro{ARMA}{Autoregressive Moving Average}
    \acro{AutoML}{Automated Machine Learning}
    \acro{avg-wQL}{Average Weighted Quantile Loss}
    \acro{CNN}{Convolutional Neural Network}
    \acro{CRPS}{Continuous Ranked Probability Score}
    \acro{ETS}{Exponential Smoothing Algorithm}
    \acro{DeepAR}{Autoregressive Recurrent Networks}
    \acro{GS}{Grid Search}
    \acro{HPO}{Hyperparameter Optimisation}
    \acro{IID}{Independent and Identically Distributed}
    \acro{LSTM}{Long Short-Term Memory}
    \acro{MAE}{Mean Absolute Error}
    \acro{MAPE}{Mean Absolute Percentage Error}
    \acro{MASE}{Mean Absolute Scaled Error}
    \acro{MLE}{Maximum Likelihood Estimation}
    \acro{MLP}{Multi Layer Perceptron}
	\acro{MQ-CNN}{Multi-Horizon Quantile Convolutional Neural Network}
	\acro{MQ-RNN}{Multi-horizon Quantile Recurrent Forecaster }
	\acro{NPTS}{Non-Parametric Time Series Algorithm}
	\acro{PACF}{Partial Autocorrelation Function}
	\acro{RNN}{Recurrent Neural Network}
	\acro{RS}{Random Search}
	\acro{Seq2Seq}{Sequence to Sequence}
	\acro{WAPE}{Weighted Absolute Percentage Error}
	\acro{wQL}{Weighted Quantile Loss}
	\acro{wQL-10}{0.1-quantile-\ac{wQL}}
	\acro{wQL-50}{0.5-quantile-\ac{wQL}}
	\acro{wQL-90}{0.9-quantile-\ac{wQL}}
\end{acronym}

\begin{abstract}
    A forecasting ensemble consisting of a diverse range of estimators for both local and global univariate forecasting, in particular \acs{MQ-CNN} \cite{wen_multi-horizon_2018}, \acs{DeepAR} \cite{salinas_deepar_2020}, Prophet \cite{taylor_forecasting_2017}, \acs{NPTS}, \acs{ARIMA} and \acs{ETS} \cite{brown_fundamental_1961}, can be used to make forecasts for a variety of problems. This paper delves into the aspect of adding different hyperparameter optimization strategies to the deep learning models in such a setup (\acs{DeepAR} and \acs{MQ-CNN}), exploring the trade-off between added training cost and the increase in accuracy for different configurations. It shows that in such a setup, adding hyperparameter optimization can lead to performance improvements, with the final setup having a 9.9 \% percent accuracy improvement  with respect to the \acs{avg-wQL} over the baseline ensemble without \ac{HPO}, accompanied by a 65.8 \% increase in end-to-end ensemble latency. This improvement is based on an empirical analysis of combining the ensemble pipeline with different tuning strategies, namely Bayesian Optimisation and Hyperband and different configurations of those strategies. In the final configuration, the proposed combination of ensemble learning and \acs{HPO} outperforms the state of the art commercial \acs{AutoML} forecasting solution, Amazon Forecast, with a 3.5 \% lower error and 16.0 \% lower end-to-end ensemble latency.

\end{abstract}

% keywords can be removed

\section{Introduction}
\subsection{Motivation}

There is a wide range of possible applications for forecasting algorithms, making it hard to find a single model that performs well on all of them. Due to this uncertainty over which models will perform best, it is common place in the forecasting space that domain experts and data scientists have to experiment with several methods before they find one that works acceptably well on a particular problem. This exploration process can be time and resource consuming and is not always practical, due to the plethora of unsolved forecasting problems as well as the scarcity of domain experts and data scientists.
\par
In recent years, \ac{AutoML} has become more popular, allowing non-technical users to solve machine learning problems without in depth knowledge about the underlying methodology, filling in for the lack of available data scientists through automation \cite{karmaker_santu_automl_2021}. In forecasting there are several approaches to AutoML, one of them being the established method of using ensemble learning and aggregation of forecasts \cite{araujo_ensemble_2007}. This has seen a recent increase in attention, with the top performing models in the M4 Competition \cite{makridakis_m4_2020} being of this nature \cite{petropoulos_forecasting_2022}.
\newline
Ensembling, can be conceptualised as the automation of the previously manual step of exploring the performance of various algorithms on a given problem and selecting the best one or a combination of models. This, however, does not address another important aspect of data science which is the selection of good hyperparameters, leading to better performance of a model trained using a particular algorithm. The combination of ensemble learning and hyperparameter tuning in an \ac{AutoML} forecasting setup will be discussed in this paper.
\par
Specifically, the paper concerns itself with the question whether hyperparameter optimisation adds accuracy improvements to a forecast ensemble (as described in \autoref{chap:experiments}) and if the corresponding latency increase is justifiable from a business perspective. It further delves into the evaluation of different tuning strategies, such as Bayesian Optimisation and Hyperband, in different variations to find an optimal configuration.
\par
To provide the necessary background for answering the scientific question, the paper will first outline the methods used as part of the research underlying this paper, focusing on the forecasting algorithms that are part of the ensemble, the different hyperparameter tuning strategies that were tested and finally, the ensemble strategy employed. The experiments that are part of this paper, combine the described methods into an end-to-end \ac{AutoML} pipeline for forecasting, that can be parameterised to run the necessary configurations for different experiments. This setup is outlined in \autoref{sec:expensemble}. To ensure more rigorous results, all experiments were repeated on four different datasets, namely Covid Death \cite{dong_interactive_2020}, Solar \cite{zhang_solar_2006}, Electricity \cite{trindade_electricityloaddiagrams20112014_2015} and Kaggle retail \cite{cook_store_2021}, the properties of which and why they were chosen is outlined in \autoref{chap:datasets}.
\newline
Chapter \ref{chap:experiments} is the core of this paper, both analysing and evaluating experiments and their findings. It first explores the accuracy and latency effect of adding hyperparameter optimisation to individual, deep-learning-based algorithms, which is discussed in \autoref{sec:expindalgs}. With the knowledge gained from these first exploratory steps, \autoref{exp:compindensemble} will detail how the changes in performance of these individual algorithms translates to the effects on the output of the ensemble. A direct comparison between the two different tuning strategies, Hyperband and Bayesian Optimisation, is part of \autoref{exp:hyperbandvsbayesian} and a more detailed analysis of the winning strategy (Hyperband) is part of \autoref{exp:strategyvariations}, which will compare different variations of set strategy as part of the ensemble. 

\subsection{Background}

This section contains a general introduction to forecasting and provides an overview of related work on the subject of \ac{AutoML} ensemble forecasting.
\par
At the root of every forecasting problem lies the intent to gather intelligence about the occurrence of a future event. In the context of time series forecasting, this means generating a prediction about the behaviour of events in the future, building on the premise that one can use past information to make inferences about the future.
Machine Learning and Statistics make use of (learned) models to find patterns in time series data and use them to generate a forecast about the future.
\newline
Forecasting problems can be differentiated as univariate and multivariate. In the former, the model used considers the past observations of a single time series to build a forecast for it. Whereas in multivariate forecasting, the model uses past observations of multiple time series to forecast those same time series \cite{januschowski_criteria_2022}. 
\newline
Furthermore, forecasting problems can be divided into local and global ones. Where the previous distinction between univariate and multivariate aimed at differentiating based on the number of time series utilised at inference time, the difference between local and global problems is determined during model formation. The parameters $\mathbf \theta$ of a local model are estimated on the target time series, meaning the time series one wants to forecast. This approach works well if a sufficient number of past observations in the target time series exists, to train the model. If this is not the case, it makes sense to consider the assumption that similar or related time series could improve performance. This is then called a global model, for which the parameters $\theta$ are estimated on a body of related time series (as well as the target time series) \cite{montero-manso_principles_2021}.
\newline
The two aforementioned categorisations aim at distinguishing based on the variables used. To introduce the aspect of certainty into the resulting forecast, one has to differentiate between point-forecasts and probabilistic models. A point forecast outputs a single value estimate for each predicted time step, according to a given metric. Whilst this makes it easier to interpret the forecast, it fails to give a sense of certainty about the prediction. Probabilistic forecasting models yield a probability distribution for each time step which provides a quantifiable way of measuring the uncertainty of the model output. These different classes of forecasting problems can be combined, leading to the categorisation in Table \ref{tab:forecast-cat}. 

% Note: exchanged h from the reference with one, assuming a single-step forecasting model.
\begin{table}[htbp]
    \centering
    \resizebox{\textwidth}{!}{%
    \begin{tabular}{ c|c|c }
        %  Note this section contains Co variate features in the Formulas:
        \hline
        \textbf{Forecast Type} & \textbf{Model Type} & \textbf{Formulation}\\
        \hline
        \multirow{4}{3cm}{Point} 
            & Local univariate 
            & $\hat \mathbf{z}_{i,t+1:t+1} = \Psi (\mathbf z_{i,1:t}, \mathbf X_{i,1:t+1})$ \\
        \cline{2-3}
        & Global univariate 
            & $\hat \mathbf{z}_{i,t+1:t+1} = \Psi (\mathbf z_{i,1:t}, \mathbf X_{i,1:t+1}, \Phi)$ \\
        \cline{2-3}
        & Local Multivariate 
            & $\hat \mathbf{Z}_{i,t+1:t+1} = \Psi (\mathbf Z_{i,1:t}, \mathbf X_{i,1:t+1})$ \\
        \cline{2-3}
        & Global Multivariate 
            & $\hat \mathbf{Z}_{i,t+1:t+1} = \Psi (\mathbf Z_{i,1:t}, \mathbf X_{i,1:t+1}, \Phi)$ \\
        \hline
        
        \multirow{4}{3cm}{Probabilistic} & Local univariate 
            & $P(\mathbf z_{i,t+1:t+1}|\mathbf z_{i,1:t}, \mathbf X_{i,1:t+1};\theta_i)$, \newline
            $\theta_i = \Psi (\mathbf z_{i,1:t}, \mathbf X_{i,1:t+1})$ \\
        \cline{2-3}
        & Global univariate 
            & $P(\mathbf z_{i,t+1:t+1}|\mathbf z_{1:t}, \mathbf X_{1:t+1};\theta_i)$, \newline
            $\theta_i = \Psi (\mathbf z_{i,1:t}, \mathbf X_{i,1:t+1}, \Phi)$ \\
        \cline{2-3}
        & Local Multivariate 
            & $P(\mathbf z_{i,t+1:t+1}|\mathbf Z_{1:t}, \mathbf X_{1:t+1};\theta_i)$, \newline
            $\theta_i = \Psi (\mathbf z_{i,1:t}, \mathbf X_{i,1:t+1})$ \\
        \cline{2-3}
        & Global Multivariate 
            & $P(\mathbf Z_{t+1:t+1}|\mathbf Z_{1:t}, \mathbf X_{1:t+1};\theta)$, \newline
            $\theta = \Psi (\mathbf Z_{1:t}, \mathbf X_{1:t+1}, \Phi)$ \\
        \hline
    \end{tabular}}
    Let $\mathbf Z$ be a set of $n$ univariate time series, where $\mathbf z_{i,1:T_i}$ is the $i$th series, with observations $z_{i,1}$ to $z_{i,T_i}$. $\mathbf Z_{t_1:t_2}$ are the values of all $n$ time series for the time slice $[t_1,t_2]$. The associated set of exogenous covariate vectors is denoted by $\mathbf X$. These can include related both dynamic and static features. $\Psi$ maps input features to a probabilistic model, parameterised by $\theta$. The parameters learned from the set of input time series are described by $\Phi$.
    \caption{Forecast Categorisation Overview based on \cite{benidis_deep_2023}}
    \label{tab:forecast-cat}
\end{table}

Ensemble learning can then be used to combine multiple forecasting algorithms to make a single prediction. Ideally, this should be done with a set of algorithms that have diverse strengths (and weaknesses), following the notion that a good ensemble strategy can determine weak spots of one algorithm and cover for them with the strengths of another one. This is not possible when very similar algorithms are chosen, which all have the same strengths and weaknesses. 
\newline
This paper uses an ensemble based on both local and global, statistical and deep-learning-based models and proposes adding hyperparameter optimisation to the latter ones to improve ensemble accuracy. To the best of my knowledge, this is the first work proposing the specific combination of ensemble learning and forecasting described in \autoref{chap:experiments}.
\par
However, similar work has already been conducted as part of several papers, such as \cite{deng_efficient_2022}\cite{jati_hierarchy-guided_2022} and \cite{abdallah_autoforecast_2022}. The former suggest a novel approach for the joint optimisation of neural architecture and hyperparameters of the entire data processing pipeline. In particular, they propose the use of Bayesian Optimisation to jointly optimize different deep-learning-based architectures as well as other pipeline configuration hyperparameters. This technique allowed them to beat multiple well-known traditional statistical models and modern deep-learning models. Their ensemble does, however, only include a set of deep-learning-based models, disregarding possible advantages of traditional statistical forecasting models like \ac{ARIMA} or \ac{ETS}. Furthermore, they only compare the accuracy of the resulting model, disregarding the practical importance of latency. This may be relevant from a business perspective, since a large search space, as suggested by \cite{deng_efficient_2022} leads to high latency. 
\newline
A different approach for \ac{AutoML} forecasting is suggested by \cite{jati_hierarchy-guided_2022}, who propose utilising hierarchy-guided model selection, addressing the problem of over fitting, in the context of time series cross validation (sequentially splitting a time series into training and validation partitions). To be precise, they leverage the hierarchical nature which often occurs in time series, by tuning hyperparameters for low level models at a higher level in the data hierarchy. A comprehensive example of this could be the problem of forecasting demand for deserts in the cafeteria of the Cooperative State University Baden-Wuerttemberg Mannheim, with sparse time series (items) for each individual type of desert on a daily basis. In the setting proposed by \cite{jati_hierarchy-guided_2022}, hyperparameters for models on the desert type basis (lowest level of the data hierarchy) would be tuned on an aggregation of these time series (e.g. by aggregating to a weekly frequency, aggregating to overall desert demand on a daily basis, or a combination of the two), allowing for more general hyperparameters and a potentially better prediction. 
\newline
Yet another variation is proposed by \cite{abdallah_autoforecast_2022}. They formulate the problem of selecting a good forecasting method as a meta-learning problem, with the goal of selecting a single best model in a way that minimizes the extensive evaluation effort associated with training a multitude of models as done in \ac{HPO}. As the approach used by \cite{abdallah_autoforecast_2022} is different from the techniques used in this paper, their methods will not be further discussed here.

\subsection{Contributions}

This Paper proposes the use of \ac{HPO} in an elaborate \ac{AutoML} ensemble as outlined in \autoref{chap:experiments}. It provides a detailed empirical analysis and evaluation of the impact of different tuning strategies and configurations on the ensemble pipeline. The experiments conducted as part of this research led to the following primary contributions:

\begin{itemize}
    \item The paper shows that the Hyperband tuning strategy is generally preferable over \break Bayesian Optimisation in a \ac{AutoML} forecast ensemble such as the one used in this paper, with Hyperband on average leading to both a 3.1 \% lower error and a 32.7 \% lower \ac{HPO} latency compared to Bayesian Optimisation. A detailed explanation of the experiment design and results can be found in \autoref{exp:hyperbandvsbayesian}.
    
    \item Results from the experiments in \autoref{exp:strategyvariations} show that optimising the configuration of the Hyperband hyperparameter tuning strategy can lead to further improvements in accuracy up to 9.9 \% over the baseline without \ac{HPO}. The paper points out that this improvement in accuracy is associated with a trade-off between error and latency. It provides an analysis of different variations of such a trade-off, pointing out good configurations for a bias toward low latency as well as low error. 
    
    \item Lastly, the paper suggests a way of finding the optimal strategy and configuration for any given bias between latency and error in \autoref{exp:strategyvariations}. It proposes tools for both a visual (\autoref{fig:thetatradeoff}) and an analytical method (\autoref{for:optimalconfig}) of determining such a configuration for a \ac{AutoML} forecast ensemble.  
    
    \item Using the optimal strategy and configurations (with the trade-off biased toward accuracy, rather than latency), namely Hyperband with a total number of 30 training jobs for tuning, the \ac{AutoML} forecast ensemble pipeline outperforms the state of the art commercial \ac{AutoML} forecasting solution Amazon forecast \cite{noauthor_time_nodate}, with a 3.5 \% lower error and 16.0 \% lower latency.
\end{itemize}

The paper also includes a comprehensive introduction to forecasting, detailing basic terminology. It further introduces all components of the custom \ac{AutoML} forecast ensemble used. This includes the forecasting algorithms, an introduction to \ac{HPO} with a focus on Bayesian Optimisation and Hyperband, as well as an overview of how ensemble learning can be used in the context of forecasting. The paper also outlines the workings of the custom forecast ensemble used in \autoref{chap:experiments}.

\section{Methodology}
\subsection{Forecasting Algorithms}
\label{sec:forecastingalgs}

The empirical analysis in this paper is based on six forecasting methods. In particular \ac{ARIMA} \cite{shumway_arima_2017}, \ac{ETS} \cite{brown_fundamental_1961}, \ac{NPTS} \cite{hardle_chapter_1994}, Prophet \cite{taylor_forecasting_2017}, \ac{MQ-CNN} \cite{wen_multi-horizon_2018} and \ac{DeepAR} \cite{salinas_deepar_2020}. This chapter will outline the way these algorithms work, assuming existing knowledge of the fundamental principles of machine learning. 

\textbf{\ac{ARIMA}} is a generalisation of the statistical forecasting model \ac{ARMA}, which itself is comprised of an autoregressive and a moving average model. Statistical autoregressive (AR) models are based on the assumption that future values can be modeled as a linear combination of past values of the same time series (from Ancient Greek autós means self). As such, an autoregressor is a point local univariate model which depends on the time series $z$ being stationary. This means that $E(\mathbf z)$ and $Var(\mathbf z)$ are constant and $\mathbf z$ has no seasonality (predictable, periodic behaviour over time).  The autoregressor with order $p$, $AR(p)$ is given by \autoref{for:autoregressor}.

\begin{equation}
 \hat z_{t+1} = \theta_0 + \theta_1 z_t + \theta_2 z_{t-1} + ... + \theta_p z_{t-p+1}
 \label{for:autoregressor}
\end{equation}

The order $p$ determines the number of lags (previous time steps taken into account) and the parameter vector $\boldsymbol{\theta}$ can be derived from the \ac{PACF} which determines the correlation of $z_{t+1}$ with $z_{t-r} \forall r \in [0,p]$, adjusted for all time steps between $t+1$ and $t-r$ ($z_{t:t-r+1}$). This means that one only cares about the direct effect of $z_{t-r}$ on $z_{t+1}$.
\newline
The second part of the \ac{ARMA} is a moving average model which is also a point local univariate model. It assumes stationary as well and is a function of the constant mean $\mu$ of $\mathbf z$ and the error of past observations. The estimator for a moving average model of order $q$ is given by \autoref{for:movingaverage}.

\begin{equation}
 \hat z_{t+1} = \mu + \theta_0 \epsilon_t + \theta_1 \epsilon_{t-1} + ... + \theta_q \epsilon_{t-q}
 \label{for:movingaverage}
\end{equation}

With $\epsilon_{i}$ being the error between the estimate $\hat z_i$ of a past observation and its actual value $\hat z_i$, $q$ determining the number of past observations taken into account ($\epsilon_{t:t-q}$), and $\boldsymbol{\theta}$ being the associated weight vector. The intuition behind the moving average is that if the scalar product of $\boldsymbol{\theta}$ and $\boldsymbol{\epsilon}$, given by $\boldsymbol{\theta}\cdot\boldsymbol{\epsilon}$ is positive, this means that the values for the past $q$ time steps $t$ were to low, and so the prediction is increased above the mean, by $\boldsymbol{\theta}\cdot\boldsymbol{\epsilon}$.
\newline
The \ac{ARMA} model is a linear combination of an autoregressive and a moving average model. As such, it also requires stationarity. The formula for an \ac{ARMA} model with arguments p and q, \ac{ARMA}(p,q) is given by \autoref{for:arma}.

\begin{equation}
 \hat z_{t+1} = \theta_0 + \theta_1 z_t + ... + \theta_p z_{t-p+1} + \mu + \theta_{p+1} \epsilon_t + ... + \theta_{p+q+1} \epsilon_{t-q}
 \label{for:arma}
\end{equation}

One of the disadvantages of \ac{ARMA} is that it requires a constant mean. \ac{ARIMA} is a generalization of \ac{ARMA} that also supports time series with a trend ($E(\mathbf z) \not = \mu$). This is achieved by including a transform from a time series $\mathbf{z}$ with a trend,  to a stationary time series $\mathbf{\tilde z}$. This transform is called integration, hence the 'I' in \ac{ARIMA}. One order of this integration is computed by $\tilde z^{(1)}_{t} = z_{t+1}-z_t$. To achieve the general case, a third argument, $d$ is introduced, which determines the order of the transform. This is best understood by considering a concrete example: If the order $d$ equals 2, then the transform is applied again to $\tilde z^{(1)}_{t}$ yielding $\tilde z^{(2)}_{t}$. For the order $d$ this is repeated recursively, resulting in $\tilde z^{(d)}_{t}$, which is stationary. \ac{ARMA} is applied to $\tilde z^{(d)}_{t}$ and the reverse of the transform is then computed for the resulting prediction, using $\hat z_{t+1} = \sum^{k-l}_{i=0} (z^{(d)}_{t-i}) + z_{t}$. For further details about the \ac{ARIMA} model, refer to \cite{shumway_arima_2017}.

\textbf{\ac{ETS}} as introduced in \cite{brown_fundamental_1961} is a local univariate time series forecasting method which is based on the notion that a future time step can be estimated as a weighted sum of past observations using smoothing. This requires the time series $\mathbf z$ to be stationary. In opposition to regression where one uses all past values at once to fit a model, smoothing recursively computes intermediate predictions, using \autoref{for:ets}. In \ac{ETS} this leads to geometrically decreasing weights, assigning lesser importance to older observations, as described in \autoref{for:ets}.

\begin{equation}
    \hat z_{t+1} = \hat z_{t} + \theta (z_{t} - \hat z_{t})
    \label{for:ets}
\end{equation}

With $\hat z_{1} = z_{0}$ and $\theta \in [0,1]$. For $\theta \rightarrow 0$ the model puts a high emphasis on historical values, in particular $\hat z_{t+1} = z_{0} | \theta = 0$ and for $\theta \rightarrow 1$ the model puts a higher emphasis on recent observation, with the extreme of $\hat z_{t+1} = z_{t} | \theta = 1$. As discussed by \cite{gardner_jr_exponential_1985}, there also exist variation of \ac{ETS} which adjust for trend and seasonality.

\textbf{\ac{NPTS}}: Standard parametric methods estimate the parameters of a specified distribution, for example using \ac{MLE} \cite{le_cam_maximum_1990}. These methods may not be efficient and/or consistent when the underlying parametric model does not match the distribution of the data. Contrastingly, non-parametric methods in forecasting have the aim of estimating quantities future time steps such as moments or density directly from past observations without fitting the parameters of a fixed distribution \cite{hardle_chapter_1994}. \ac{NPTS} method used predicts a probability distribution for future time steps, for each time series individually, making it a probabilistic local univariate model. The statistical estimation is based on a sample taken from past observations, which can be done with different linear and non-linear methods, the former can be expressed using a kernel function and will be the focus of this section. One can further distinguish non-parametric methods for forecasting into autoregressive problems and regression with
correlated errors. The former refers to the prediction of a future time step $z_{t+1}$ based on past observations of $\mathbf{z}$ and the latter to predicting $z_{t+1}$ based on observations of the exogenous variable  $\mathbf{x}$. The autoregressive estimation of the first and second moment is given by Formula \ref{for:nonparmoments}.

\begin{equation}
    \hat{\mu}_{\mathbf{z},t+1} = E(\mathbf{z}_{t+1} | \mathbf{z}_{1:t}),\;\;\; 
    \hat{\sigma}^2_{\mathbf{z},t+1} = Var(\mathbf{z}_{t+1} | \mathbf{z}_{1:t})
    \label{for:nonparmoments}
\end{equation}

The experiments described in \autoref{chap:experiments} use a kernel with exponential weight decay. This means that observations closer to the current time step are sampled with a higher probability than those further in the past. The kernel $K$ is given by $K(u)=\alpha \cdot e^{-\lambda \cdot |u|}$, where $\alpha$ is the scaling factor, controlling that $\int K(u)\, du = 1$ and $\lambda$ being the influencing the rate of weight decay. The explicit density estimator for this case is given by \autoref{for:nonparexpkernelestimator}.

\begin{equation}
    \hat{f}(z_{t+1} | {t+1}) = \frac{1}{th} \sum_{i=1}^{t} K\left(\frac{y_{t+1} - y_i}{h}\right)
    \label{for:nonparexpkernelestimator}
\end{equation}

With $t$ being the number of historical observations in the time series and $h$ the bandwidth parameter which controls the width of the kernel window. 

\textbf{Prophet:} The prophet model, as introduced by \cite{taylor_forecasting_2017} is a probabilistic univariate curve fitting model, that aims at providing a flexible and interpretable basis for forecasting. Prophet consists of three major components: the trend function $g(t)$ representing non-periodic changes over time, seasonality $s(t)$ modeling periodic behaviour of different frequencies, and holidays $h(t)$ for irregularly occurring events. \newline
The trend component $g(t)$ contains two variations: a saturating growth model, and a piecewise linear model.  The latter is used for modeling and forecasting saturated growth as it typically occurs for a population in natural ecosystems. A simple form of $g(t)$ as a logistic growth model is given by $g(t) = \frac{C}{1+\exp(-k\cdot(t-m))}$, with $C$ being the carrying capacity (meaning the value toward which the growth converges), $k$ being the growth rate and $m$ the offset parameter. To make this approximation adaptable to a changing carrying capacity and alternating growth rate, \cite{taylor_forecasting_2017} suggest making both of them a function of time leading to \autoref{for:prophetsatgrowth}.

\begin{equation}
    g(t) = \frac{C(t)}{1+\exp(-(k + \mathbf a(t)^T\boldsymbol \delta)\cdot(t-(m +\mathbf a(t)^T \boldsymbol \gamma)))}
    \label{for:prophetsatgrowth}
\end{equation}

Here $C(t)$ denotes the carrying capacity dependent on time and $\boldsymbol \delta \in \mathbb{R}^S$ being a vector of $S$ growth rate adjustments. These changepoints are incrementally activated by computing the scalar of $\boldsymbol \delta$ and $\mathbf a(t) \in\{0,1\}^S$, with $\mathbf a_j(t) = 1$ if $t\le s_j$. The changepoints $s_j$ can be either manually or automatically selected. Furthermore, the offset $m$ must be adjusted to control for the changing growth rate. This is done by replacing $m$ with $m +\mathbf a(t)^T \boldsymbol \gamma$, where $\boldsymbol \gamma$ is given by \autoref{for:prophetsatgrowthgamma}.

\begin{equation}
    \gamma_j = \Big( s_j - m - \sum_{l<j} \gamma_l \Big) \Big( 1- \frac{k+\sum_{l<j}\delta_l}{k+\sum_{l\le j}\delta_l}\Big)
    \label{for:prophetsatgrowthgamma}
\end{equation}

The piecewise linear model can be generally used to represent growth where it is not saturated. It can be conceptualised as breaking a non-linear problem down into multiple stages, so that each can be linearly modeled. These stages are separated by breakpoints which are represented using the growth rate adjustment vector $\mathbf a_j(t)$. This leads to $g(t) = (k + \mathbf a(t)^T \boldsymbol \delta)t + (m + \mathbf a(t)^T \boldsymbol \gamma)$, where $\gamma_j = s_j\delta$, yields a continuous function.
For both trend models, the existence of future changepoints can be simulated based on information from past changepoints.
\par
The seasonality in Prophet, which models periodically reoccurring patterns in the time series with different frequencies, is realized using a standard Fourier series \cite{harvey_structural_1993}. The concept of which is, to use the sum of trigonometric functions to approximate a well-behaved function. Hence, the seasonal component $s(t)$ is given by $\sum_{n=1}^N(a_n cos (\frac{2\pi nt}{P})+b_n cos (\frac{2\pi nt}{P}))$ and parameterized by $\boldsymbol{\theta} = \{a_1,b_1,...,a_N,b_N\}$.
\par
Holidays and irregularly occurring events $i$ are modeled using a vector of indicator functions $Z(t)$ (one for each Holiday) that returns $1$ if $t$ is an element of the set of past and future dates of the event $D_i$ ($Z(t) = [\mathbf 1_{t\in D_1},..,\mathbf 1_{t\in D_L}]$). The vector $\boldsymbol \kappa$ contains the the changes to the forecast for the respective holidays, with $\kappa_i$ being the forecast shift for event $i$, such that $h(t) = Z(t)\boldsymbol{\kappa}$. 
\newline
The final forecast is then computed as the sum of trend, seasonality and events, as in \autoref{for:prophet}, using posterior sampling making the forecast probabilistic.

\begin{equation}
    \hat z(t+1) = g(t+1)+s(t+1)+h(t+1)
    \label{for:prophet}
\end{equation}

\textbf{\ac{MQ-CNN}} as introduced by \cite{wen_multi-horizon_2018}, is a probabilistic forecasting model which is trained on a set of time series to predict a single target time series, making it a global univariate model. Furthermore, it is a \ac{Seq2Seq} model based on a \ac{CNN} encoder and a \ac{MLP} decoder, yielding Multi-horizon Quantile forecasts. These concepts are detailed in the following.
\newline
A \ac{Seq2Seq} model is a model which takes a sequence such as natural language or a time series as an input to compute an outputs sequence. Some of the previously outlined models, namely \ac{ARIMA}, \ac{ETS} and \ac{NPTS} interactively forecast a single next value $z_{t+1}$. They do this recursively until a forecast with the length $K$ of the desired forecasting horizon is generated, which leads to the accumulation errors from previous single steps. However, in Muli-horizon forecasting, the model directly predicts multiple steps $(z_{t+1},...,z_{t+k})$. As suggested by \cite{taieb_bias_2016} this leads to avoidance of error accumulation whilst retaining efficiency through sharing parameters, making it preferable over single step forecasting. 
\newline
Like \ac{NPTS}, Quantile regression is a non-parametric method for capturing the distribution of a time series. It estimates the conditional quantiles of the target $P(z_{t+k}<z^{(q)}_{t+k}|z_{1:t})$, leading to a robust, accurate forecast with sharp prediction intervals \cite{taylor_quantile_1999}. This type of forecast, produces an output matrix $\hat \mathbf{Z} = [\hat z^{(q)}_{t+k}]_{k,q} \in \mathbb{R}^{Q\times K}$ ($Q$ being the number of predicted quantiles and $K$ the number of horizons/predicted time steps). The respective model is trained using a quantile loss function, such as the total quantile loss, given by $\sum_t\sum_q\sum_k L_q(z_{t+k},\hat z^{(q)}_{t+k})$. 
\par
The primary focus of \cite{taylor_quantile_1999}  is a \ac{MQ-RNN}. However the paper also suggests that the MQ-framework works with any neural network that has a sequential or temporal structure and is compatible with forking-sequences, such as WaveNet \cite{oord_wavenet_2016}. In the experiments described in this paper, WaveNet, which is a hierarchical causal convolutional network, is used as the encoder, hence \ac{MQ-CNN}. 
This encoder outputs hidden states $h_t$ for all past observations, which then serve as input for the forked (two branch) \ac{MLP} decoder. The decoder consists of a global branch $m_G(\cdot)$ and a local branch $m_L(\cdot)$. The former takes the aforementioned hidden states and future inputs from exogenous covariate vectors $\mathbf x^{(f)}_{t:,i}\in\mathbf X$ (which is here comprised of past $\mathbf x^{(h)}_{:t,i}$, future $\mathbf x^{(f)}_{t:,i}$ and static $\mathbf x^{(s)}_{i}$ covariate features) to compute a series of horizon-specific contexts $c_{t+1},...,c_{t+K}$ as well as a horizon-agnostic context $c_a$, as formalized in \autoref{for:mqcnnmlp} (a).

\begin{equation}
    (a)\;m_G(h_t,x^{(f)}_{t:})=(c_{t+1},...,c_{t+K},c_a);\;\;(b)\;m_L(c_{t+k},c_a,x^{(f)}_{t+k}) = (\hat z^{(q_1)}_{t+k},...,\hat z^{(q_Q)}_{t+k})
    \label{for:mqcnnmlp}
\end{equation}

The local MLP is then applied for each time step that is predicted, yielding the final forecast. It computes the output quantiles for each time step based on the corresponding future input and the two contexts from the global MLP as formalized in \autoref{for:mqcnnmlp} (b).

\begin{figure}[H]
    \caption{Computation in a single \ac{LSTM} cell}
    \includegraphics[width=\textwidth]{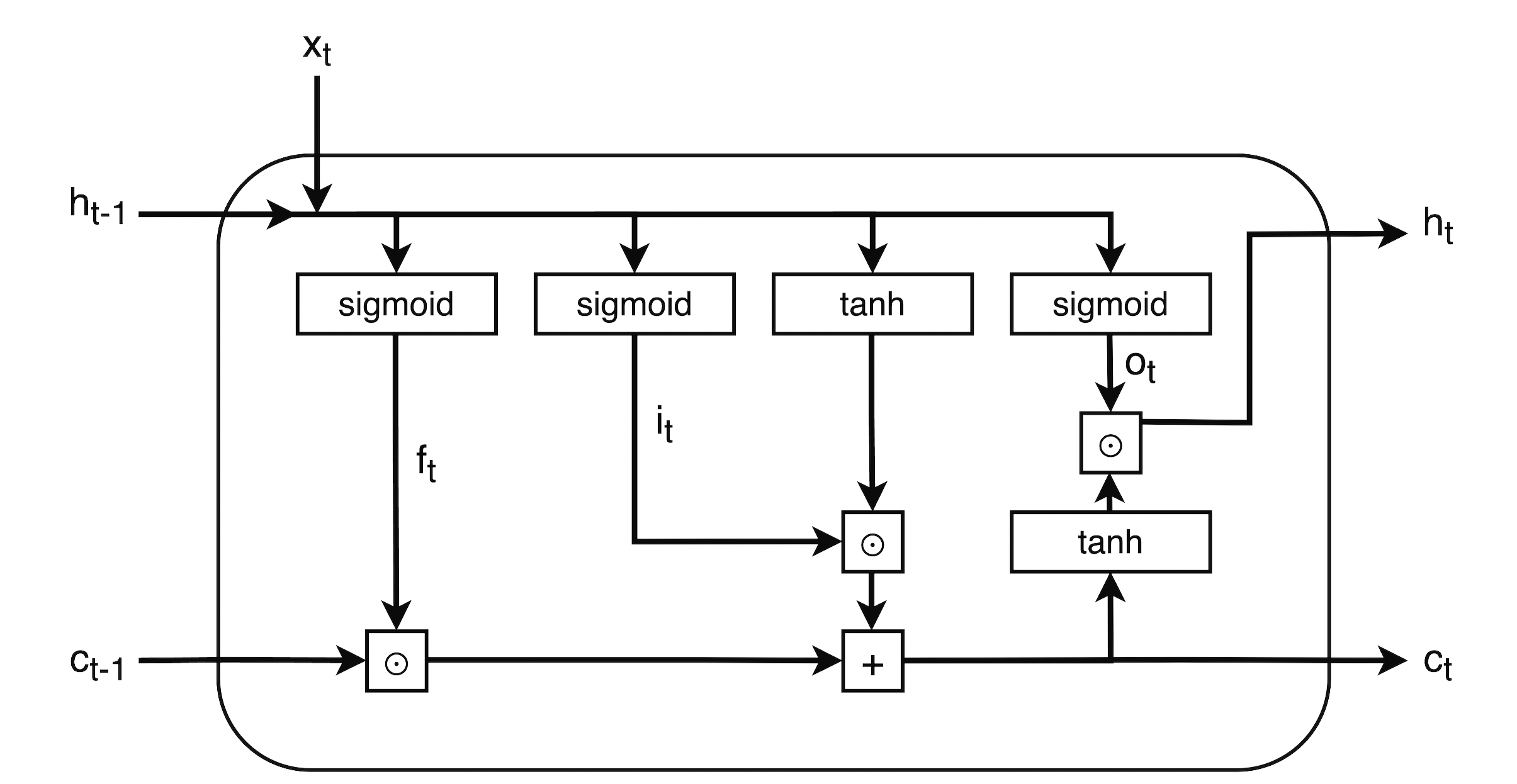}
    \label{fig:lstm}
    \centering
\end{figure}

\textbf{\ac{DeepAR}} \cite{salinas_deepar_2020} is a deep-learning-based autoregressive probabilistic global univariate forecasting model, which is based on a \ac{LSTM} model. A \ac{LSTM} \cite{hochreiter_long_1997}\cite{goodfellow_deep_2016} is a modification of a standard \ac{RNN} that deals with the vanishing or exploding gradient problem (information from some hidden state $h_n$ has a exploding or vanishing effect on a later hidden state $h_{n+m}$, depending on the weights of the network, making it difficult for the network to detect the relevant signal). It does this by introducing the cell state, which has the purpose of capturing useful long-term information, and three gates: the forget-gate, the input-gate and the output-gate, which each represent an attention mechanism (here weight matrix) that determines relevant information. The forget gate calculates a weight matrix $f_t$ with the relevance of information from the previous cell state $c_{t-1}$. The input gate calculates a similar weight matrix for the preliminary current cell state $\tilde c_t$. The current cell state $c_t$ is then computed as $f_t \odot c_{t-1} + i_t \odot \tilde c_t$ (with $\odot$ denoting element-wise matrix multiplication), essentially making $c_t$ a combination of previous and current relevant information. The output gate $o_t$ is computed to determine what information of $c_t$ is relevant to the current hidden state $h_t$, which is then calculated as a function of $o_t$ and $c_t$. The computation in a single \ac{LSTM} cell can be visualized as in \autoref{fig:lstm}.

\ac{DeepAR} uses such an \ac{LSTM} for both the conditioning (encoder) range, in which the model recursively takes in observations from past time steps, and in the prediction (decoder) range, where the model continues to unroll predicting the distribution of future time steps as given by 

\begin{equation}
    P(\mathbf z_{i,t+1:t+K}|\mathbf z_{1:t}, \mathbf x_{1:t+1}) 
    \label{for:deepar}
\end{equation}

According to \cite{salinas_deepar_2020} \ac{DeepAR} performs particularly well when provided with large amounts of related and covariate time series, it offers the advantages of minimal manual intervention by experts, due to the fact that seasonal behaviors and dependencies on given covariates across time series are automatically modelled. It further provides the ability to compute forecasts for time series with few or no past observations, as long as they are related to the time series the model was trained on.

\subsection{Algorithm Evaluation}
\label{sec:algeval}

In the empirical evaluation described in \autoref{chap:experiments} the aforementioned algorithms and the resulting ensemble are evaluated using a set of metrics, namely: \ac{MAPE}, \ac{MASE}, \ac{WAPE} and \ac{wQL}. In the following section, these metrics are described in more detail.
\newline
\textbf{\ac{MAPE}} is an average over the absolute percentage error between predicted and true values for each time step. It is particularly useful if the time series has a high variance, with outliers having a higher impact. \ac{MAPE} can be computed as shown in \autoref{for:mape}, with $n$ being the number of time steps, $z_t$ the observed value at time step $t$ and $\hat z_t$ being the forecasted value at $t$.

\begin{equation}
    MAPE =  \frac{1}{n}\sum^n_{t=1}|\frac{z_t-\hat z_t}{z_t}|
    \label{for:mape}
\end{equation}

\textbf{\ac{MASE}} is a scaled version of the \ac{MAE}. The scaling factor is the \ac{MAE} of the naive forecast method, which in the non-seasonal case, simply predicts the previous time step $\hat z_t = z_{t-1}$ and in the seasonal case predicts the time step from one season ago, $\hat z_t = z_{t-m}$ where $m$ is the seasonality value computed over the training set of length $d$. This makes \ac{MASE} well suited for periodic time series. The seasonal metric can be computed as in \autoref{for:mase}, for the non-seasonal version, simply set $m=1$.

\begin{equation}
    MASE =  \frac{1}{n}\sum^n_{t=1} \left( \frac{|z_t-\hat z_t|}{\frac{1}{d-m}\sum^d_{j=m+1}|z_j-z_{j-m}|} \right)
    \label{for:mase}
\end{equation}

\textbf{\ac{WAPE}} represents the general deviation of forecasted values from true values, calculated as the error of the sum of all forecasted values divided by the sum of all actual values, as in \autoref{for:wape}. It may be advantageous to use \ac{WAPE} when there are missing values or near zero time steps.

\begin{equation}
    WAPE =  \frac{\sum^n_{t=1}|z_t-\hat z_t|}{\sum^n_{t=1}|z_t|}
    \label{for:wape}
\end{equation}

\textbf{\ac{wQL}} is used to determine the model performance for particular quantiles. This can be utilised for making better predictions on a particular quantile, which could be used to avoid under or over predicting. This case may be useful when one is forecasting customer demand for a product and would rather have too much stock than run out and have to turn away customers, making a model that avoids under predicting favorable. 
\newline
In this paper I consider the \ac{avg-wQL}, the \ac{wQL-10}, \ac{wQL-50} and the \ac{wQL-90}. The latter three are computed using \autoref{for:wql}, with the predicted $\tau$-quantile at time step $t$ being $q^{(\tau)}_t$ and $\tau \in {0.1,0.5,0.9}$ (or in the general case $\tau \in (0,1)$).

\begin{equation}
    wQL(\tau) = 2 \frac{\sum^n_{t=1}(\tau \max(z_t-q^{(\tau)}_t, 0) + (1-\tau)\max(q^{(\tau)}_t-z_t, 0))}{\sum^n_{t=1}|z_t|}
    \label{for:wql}
\end{equation}

For any time step, only one of the two $\max(\cdot)$ terms in the numerator of \autoref{for:wql} is unequal to zero. If the estimator under predicts the left side, weighted by $\tau$ is "active" and when over-predicting the right side, weighted by $1 - \tau$ is "active". For $\tau > 0.5$ under predictions are stronger penalized than over predictions and for $\tau < 0.5$ it is the other way around. If $\tau = 0.5$ then $wQL(0.5) = WAPE$. The \ac{avg-wQL} is defined in \autoref{for:avgwql}, where the sum over $num\_quantiles$ takes the values of the desired quantiles. It approximates the \ac{CRPS}, which is a strictly proper scoring rule.

\begin{equation}
    avg\_wQL = \frac{\sum_{m=1}^{num\_quantiles}wQL(m)}{m},
    \label{for:avgwql}
\end{equation}

\subsection{Hyperparameter Optimisation}
\label{sec:hpo}

The experiments conducted as part of this paper compare different \ac{HPO} strategies. This section will provide a brief introduction to \ac{HPO} and explore the used optimisation algorithms, namely: Bayesian Optimisation \cite{shahriari_taking_2016} and  successive halving techniques, in particular Hyperband \cite{li_hyperband_2018}.
\newline
The aim of a (parametric) machine learning algorithm is to estimate the parameters that best fit the problem at hand. The machine learning algorithm that is used to learn the parameters is itself parameterised by so-called hyperparameter (e.g. learning rate or number of hidden layers in the case of a neural-network-based model). These hyperparameters are used to determine different aspects of the final model, such as its architecture and the manner in which the model is trained. Finding the hyperparameters that lead to the optimal solution of the machine learning problem is called hyperparameter tuning. This process can be performed manually, which may require detailed understanding of the model and experience with similar problems. Hyperparameter optimisation techniques have the objective of automating this process of finding the best hyperparameters. Here the user has to specify a search space, meaning ranges or options for possible hyperparameters which the \ac{HPO} strategy then automatically explores. A simple example of such a tuning technique is \ac{RS} which assumes a probability distribution (e.g.uniform) over the search space and then samples random configurations, training and evaluating a model for each.  This is similar to Grid Search \ac{GS}, which exhaustively searches for the optimal hyperparameter combination, as \ac{RS} also treats each configuration independently and simply selects the best one after the strategy is done sampling. In practice, these simple techniques can lead to high tuning latency, due to the potentially expansive search space resulting from even a few tuned hyperparameters. The techniques used in the experiment section of this paper are more sophisticated in that they exploit information from well performing configurations (Bayesian), or stop pursuing configurations which seem to perform poorly (Hyperband) and by doing so save computational resources or speed up latency.

\textbf{Hyperband} is a \ac{HPO} technique that formulates the problem of finding optimal hyperparameters as a  pure-exploration nonstochastic infinite-armed bandit problem where a predefined resource is allocated to randomly sampled configurations \cite{li_hyperband_2018}. As such, it is based on successive halving \cite{jamieson_non-stochastic_2016}, which is a technique that allows speeding up the evaluation of randomly sampled configurations, by early stopping those that perform badly. This is done by dividing the available computational budget $B$ across a fixed number $n$ of configurations. After a set measurement (e.g. number of epochs, time, etc.) the algorithms are evaluated and the worst performing halve is terminated. The freed up budget is added to the remaining training jobs, effectively doubling it. This is repeated until only a single, winning hyperparameter configuration is left \cite{hutter_automated_2019}. Given the budget $B$ and the variable number of configuration samples $n$, successive halving allocates $B/n$ resources to each configuration. This leaves $n$ as a parameter to be set by the user, which is where Hyperband comes into play. It does a \ac{RS} over the hyperparameter search space for $n$ configurations, allocating a minimum resource amount $r$ to all configurations, before some are terminated (with large $n$ corresponding to smaller $r$ and hence earlier stopping). It is refereed to as nonstochastic due to the procedural way in which resources are allocated. The way Hyperband works is formalised in Algorithm \ref{alg:hyperband},

\begin{algorithm}
\caption{Hyperband algorithm for hyperparameter optimization \cite{li_hyperband_2018}}
\label{alg:hyperband}
\floatname{algorithm}{Procedure}
\renewcommand{\algorithmicrequire}{\textbf{Input:}}
\renewcommand{\algorithmicensure}{\textbf{Initialization:}}
\begin{algorithmic}[1] 
\Require $R, \eta\; (default\; \eta=3)$
\Ensure $s_{max} = \lfloor \log_{\eta}(R)\rfloor,\; B = (s_{max}+1)R$
\For{$s\in\{s_{max},s_{max}-1,...,0\}$}
    \State $n = \lceil \frac{B}{R}\frac{\eta^s}{(s+1)}\rceil,\;\; r = R\eta^{-s}$
    \State \texttt{// begin successive halving with (n,r) inner loop}
    \State $T =$ \texttt{get\_hyperparameter\_configuration}$(n)$
    \For{$i\in\{0,...,s\}$} 
        \State $n_i = \lfloor n\eta^{-i}\rfloor$
        \State $r_i = r\eta^{i}$
        \State $L=\{$\texttt{run\_then\_return\_val\_loss}$(t,r_i):t\in T\}$
        \State $T=\{$\texttt{top\_k}$(T,L,\lfloor n_i/\eta\rfloor)$

    \EndFor
\EndFor
\State \Return configuration with the smallest loss seen so far.
\end{algorithmic}
\end{algorithm}

which consists of an outer and an inner loop. The former (lines 1-2) selects different values for $n$ and $r$ and the latter (lines 4-10) then performs an execution of successive halving with the selected $n$ and $r$, referred to as bracket. The Hyperband algorithm takes two parameters: the maximum amount of available resources $R$ and $\eta$ which controls the share of configurations terminated in each round. 
\newline
The algorithm further builds on \texttt{get\_hyperparameter\_configuration}$(n)$ to return a set of \ac{IID} sampled configurations from the distribution defined over the search space (here uniform), \texttt{run\_then\_return\_val\_loss}$(t,r_i)$ to run the respective machine learning algorithm with hyperparameter configuration $t$ on the allocated amount of resources $r$, returning their losses and \texttt{top\_k}$(T,L,\lfloor n_i/\eta\rfloor)$ to return the k best configurations given a set of configurations $T$ and their losses $L$.

\textbf{Bayesian Optimisation} is a \ac{HPO} technique which assumes a black-box function $f$ describing the algorithm performance dependent on the hyperparameter configuration $\mathbf x$ element of the search space $X$. This function $f$ is unknown but can be sampled by running and evaluating particular hyperparameter configurations.
\newline
The idea of Bayesian Optimisation is to start from a prior assumption about the objective function $f$ and continually refine it via Bayesian posterior updating based on information from sampled and evaluated configurations. It always selects the next sample configuration based on the current assumption about the objective function, exploiting information gained from previously evaluated configurations. Once that maximum number of iterations has been reached, the strategy returns its best estimate of the optimal configuration.
\newline
The initial objective function can be estimated from a set of sample configurations for which the algorithm in question is run and evaluated. The estimation of $f$ is based on data points resulting from the first evaluated configurations and can be fitted using a Gaussian process regressor. Such a regressor is defined through the mean and standard deviation of a set of multiple different regressors, indicating both the expected value and associated uncertainty for each configuration. Using the current estimate of $f$, the acquisition function $\alpha(\mathbf x)$ can then be calculated. The goal of Bayesian Optimisation is to optimize this acquisition function and then select its $arg\max$ as the next sample configuration, formalizing the trade-off between exploration and exploitation. Sampling and estimation of $f$ and calculation of the acquisition is repeated for a fixed number of iterations $N$ as described in algorithm \ref{alg:bayesianopt}.

\begin{algorithm}[htbp]
\caption{Bayesian optimization \cite{shahriari_taking_2016}}
\label{alg:bayesianopt}
\begin{algorithmic}[1] 
\For{$n\in\{1,2,...,N\}$}
    \State $\mathbf x_{n+1} = {arg\,max}_{\mathbf x} \alpha(\mathbf x;D_n)$ \Comment{select new configuration by optimizing the acquisition function}
    \State $y_{n+1} =$ \texttt{run\_algorithm}$(\mathbf{x}_{x+1})$
    \State $D_{n+1} = \{D_n,(\mathbf{x}_{x+1}, y_{n+1})\}$ \Comment{augment available data}
    \State $T =$ \texttt{get\_hyperparameter\_configuration}$(n)$
    \State update statistical model
\EndFor
\State \Return configuration with the smallest loss seen so far.
\end{algorithmic}
\end{algorithm}

\begin{figure}[htbp]
    \caption{Bayesian Optimisation Visualisation from \cite{shahriari_taking_2016}}
    \centering
    \includegraphics[width=.7\textwidth]{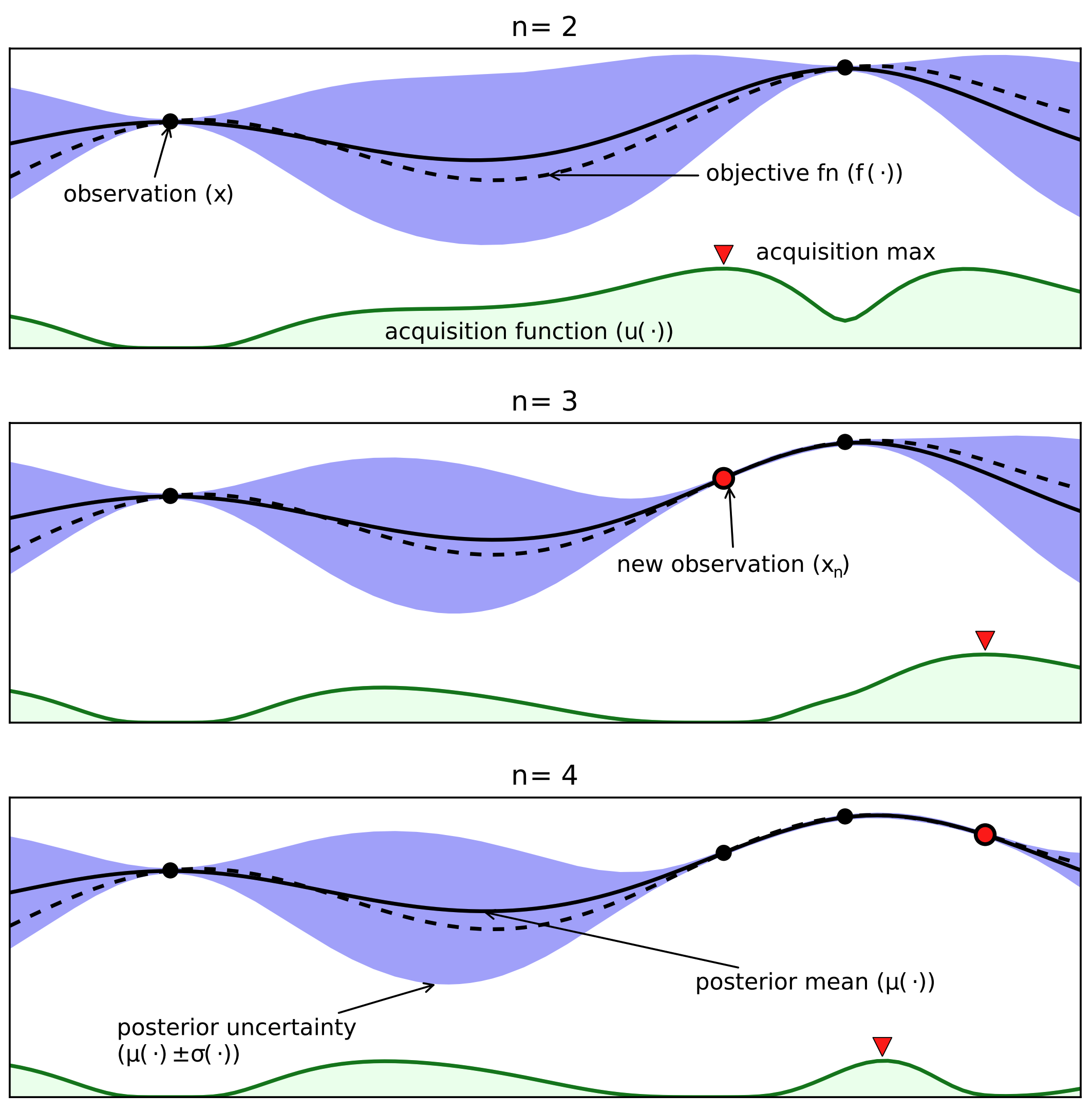}
    \label{fig:bayesianopt}
    \break
    Three iterations of Bayesian Optimisation (iterations 2-4), with the dashed line representing the objective function and the solid line representing the probabilistic surrogate model.
    \centering
\end{figure}

As a result of this approach, the Bayesian optimisation strategy uses available information $D_n$ to select the next sample. This differentiates it from the previously discussed strategies, which select new configurations independently through random sampling. 
\newline 
A simple visual example with $\mathbf{x},y\in \mathbb{R}^1$ and $N=4$ can be found in \autoref{fig:bayesianopt}. For further details on Bayesian optimisation refer to \cite{shahriari_taking_2016} and \cite{hutter_automated_2019}.

\section{Forecasting Ensemble}
\label{sec:ensemble}

In forecasting, an ensemble can be used to combine a variety of different data sources and algorithm types to provide a more accurate and robust forecast. This chapter will provide a brief introduction to ensemble strategies, in particular those used in forecasting. It will further detail the ensemble method used for the general purpose forecasting setup at the core of this paper. 
\newline
In general, the goal of ensemble learning is to use a combination of different algorithms and data augmentation to obtain multiple diverse models. These are then combined through aggregation, averaging or voting strategies, with the goal of getting a more robust and accurate prediction \cite{dong_survey_2020}. In the forecasting space, ensemble learning first originated in weather forecasting \cite{wu_ensemble_2021}, where it has led to improvements in both accuracy and robustness of forecasts. Recently, ensemble forecasting has seen an increase in attention, with the top performing models in the M4 Competition \cite{makridakis_m4_2020} being ensembles. 
\newline
When considering a set of forecasting algorithms all trained on the same data, one strategy for generating a good forecast of the resulting models is to simply use the model that performs best on the test data. Even though this approach may be intuitive and simple, it is naive in that it could make the resulting forecast overconfident if the test time series is not representative of the future. A more robust strategy could be to calculate an average over predicted values for each time step and use the resulting time series as the final prediction. To avoid having poorly performing algorithms decrease the ensemble performance, one can furthermore extend this strategy by only considering the $k$ best algorithms in the average. This can be done both on a global and a local level. In the former case, one would consider the performance across all items for selecting the $k$ best algorithms and would then use them to make inference for all of the items (time series). In the latter (local) case, one would evaluate the performance of all algorithms for each item individually and then select the $k$ best ones at an algorithm level. This can lead to different algorithms being selected for different items in the same dataset. 
\newline
The forecast ensemble setup used in this paper uses a combination of local and global strategies to compute the ensemble. This leads to a three dimensional method, with a local, a global and a combination parameter. The local parameter is used to set a threshold for how well an algorithm needs to perform compared to the best performing one, on the item level, to be considered in the local ensemble. The global parameter follows the same concept but is used to compare models across all items. The combination parameter, is then used to set a threshold on a per item basis, for which strategy to use: global or local. The idea of this ensemble family (set of ensemble strategies parameterised by local, global and combination parameter) is to find a good trade of between the advantage of the local strategy, namely item level specificity and the robustness of the global strategy. Finding optimal values for these three parameters, respectively, finding a concrete ensemble strategy, is framed as another optimisation problem on a second test window. Since the precise mechanics of the ensemble family used are not subject of the empirical investigation of this paper, I will not delve into more detail here and further assume the strategy as a given.

\section{Datasets}
\label{chap:datasets}
To ensure rigorous and verifiable results, all experiments included in this paper were conducted on a list of public open source datasets. This list includes the following four datasets: Covid Death \cite{dong_interactive_2020}, Solar \cite{zhang_solar_2006}, Electricity \cite{trindade_electricityloaddiagrams20112014_2015} and Kaggle retail \cite{cook_store_2021}. These data sets were chosen due to their diverse characteristics, in terms of number of observations, number of items and frequency, and common use in the forecasting literature.

\begin{figure}[H]
    \caption{Covid death dataset overview}
    \includegraphics[width=.9\textwidth]{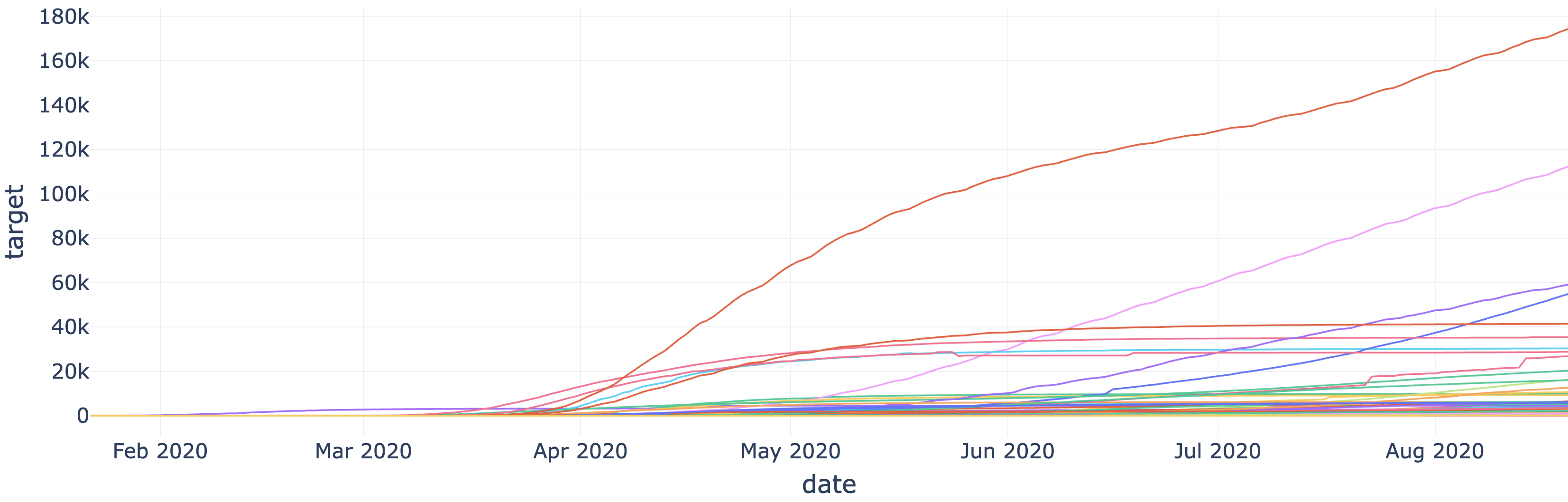}
    \label{fig:covid_death}
    \centering
\end{figure}

% TODO: Add a visual analysis of the datasets
\textbf{Covid Death:} This is a dataset provided by the Center for Systems Science and Engineering at Johns Hopkins University \cite{dong_interactive_2020}. The dataset is made up of 230 items (time series)  with 212 time steps each. It was also used for benchmarks in \cite{petropoulos_forecasting_2022}\cite{dong_interactive_2020} and contains reported coronavirus disease 2019 deaths on a daily basis, as visualized in \autoref{fig:covid_death}. The data is at the province level in China; at the city level in the USA, Australia, and Canada; and at the country level, otherwise. In the experiments, 30 days were forecasted for items in this dataset.

\begin{figure}[H]
    \caption{Solar dataset overview}
    \includegraphics[width=.9\textwidth]{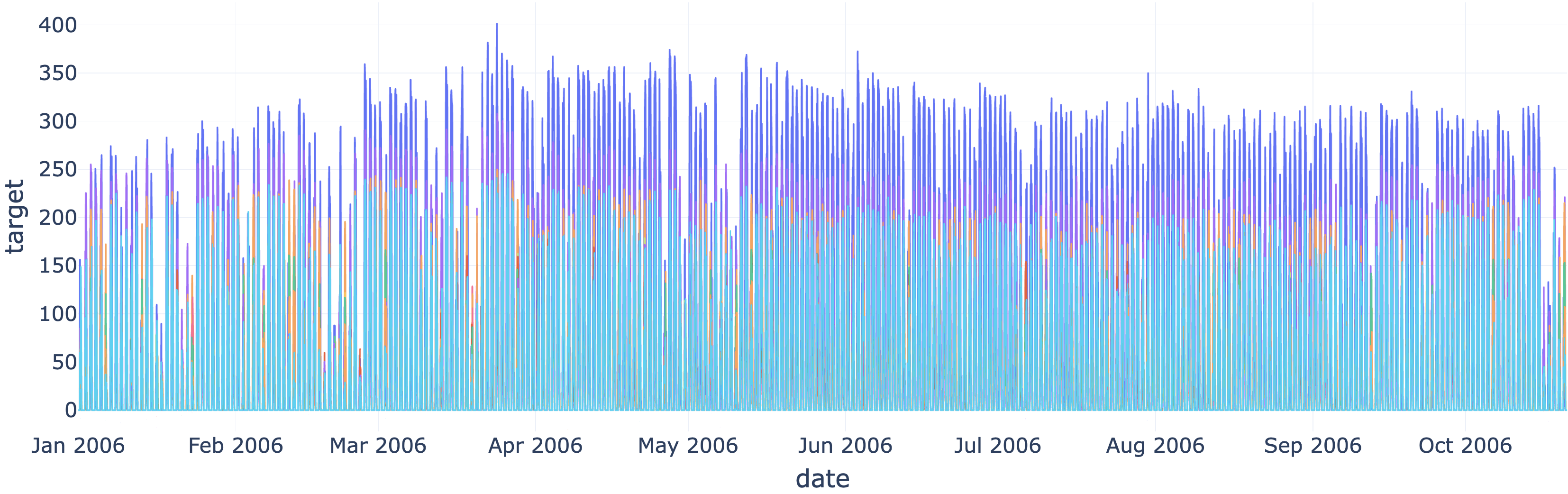}
    \label{fig:solar}
    \centering
\end{figure}

\textbf{Solar:} The Solar dataset \cite{zhang_solar_2006}, provided by the National Renewable Energy Laboratory of the U.S. Department of Energy, provides data on the energy production of photovoltaic power stations on an hourly basis, as visualized in \autoref{fig:solar}. It contains 137 items, with 7033 time steps each. In the experiments, 168 hours (7 days) were forecasted for items in this dataset.

\begin{figure}[H]
    \caption{Electricity dataset overview}
    \includegraphics[width=.9\textwidth]{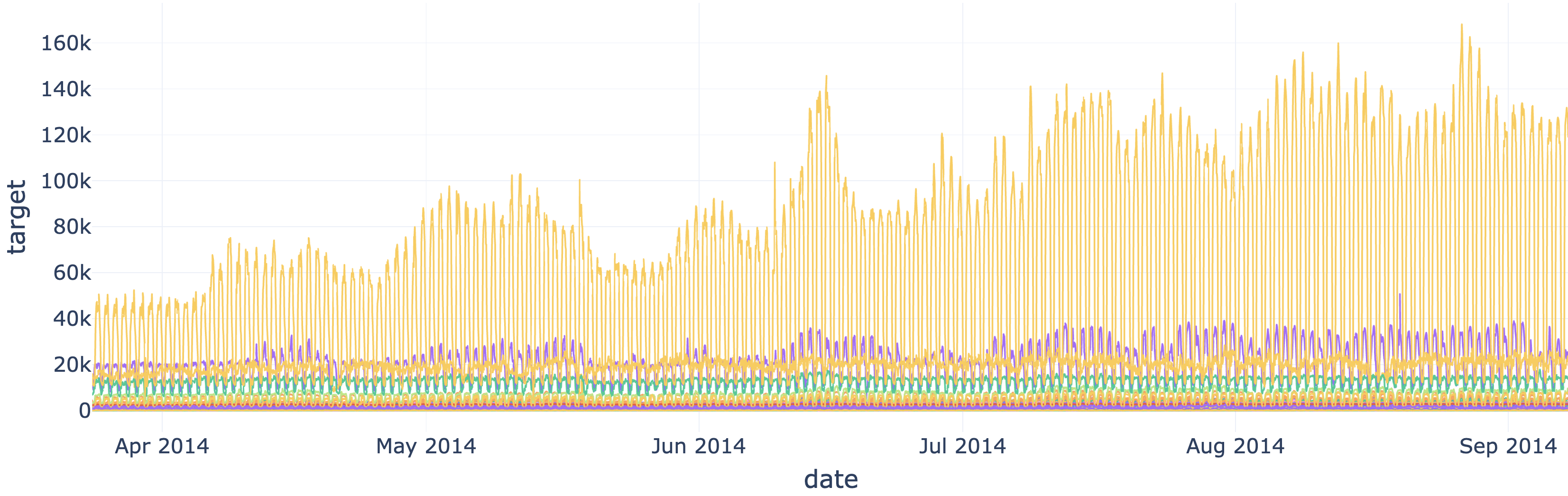}
    \label{fig:electricity}
    \centering
\end{figure}

\textbf{Electricity:} This dataset \cite{trindade_electricityloaddiagrams20112014_2015}, as also referenced by \cite{salinas_deepar_2020}\cite{yu_temporal_2016}\cite{rangapuram_deep_2018}, contains energy consumption of 370 different clients (370 items) with 4032 observations each in hourly frequency, with no missing values, as visualized in \autoref{fig:electricity}. In the experiments, 168 hours (7 days) were forecasted for items in this dataset. 

\begin{figure}[H]
    \caption{Kaggle retail dataset overview}
    \includegraphics[width=.9\textwidth]{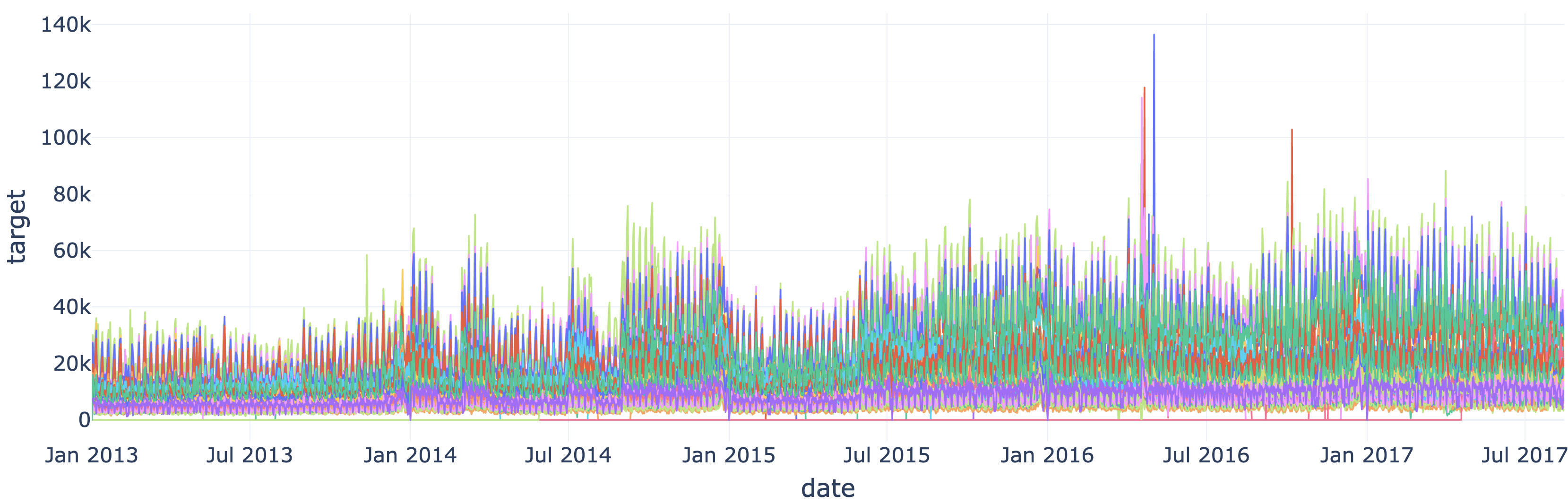}
    \label{fig:kaggleretail}
    \centering
\end{figure}

\textbf{Kaggle retail:} The Kaggle retail dataset \cite{cook_store_2021}, as used in the experiments, contains data on the number of sales of different product families at a daily frequency, as visualized in \autoref{fig:kaggleretail}. It includes 54 different items with 1684 time steps each and is used with a forecasting horizon of 14 days in the experiments.

\section{Experiments}
\label{chap:experiments}
This chapter contains the design and evaluation of results for all experiments conducted as part of this paper. Due to the relatively low number of repetitions for each experiment with different random seeds, the use of p-values may be misleading and is therefore avoided throughout the  chapter. It instead uses mean values and standard deviations for the comparison of results. 

\subsection{Impact on Individual Algorithms}
\label{sec:expindalgs}

The scientific question of this paper is based on the premise that \ac{HPO} provides accuracy improvements to single algorithms. This makes it then further reasonable to explore how this improvement affects the accuracy of a forecasting ensemble. In the following, this paper shows that the aforementioned premise is true for the deep learning algorithms in question. To be precise, this section explores the setup and results of experiments which quantify the accuracy and cost changes from adding HPO to individual algorithms, namely \ac{DeepAR} and \ac{MQ-CNN}. 
The setup compares the results of two different machine learning pipelines: \texttt{pipeline\_1} and \texttt{pipeline\_2}. In machine learning, a pipeline is used to automate workflows, including data preprocessing and transformation as well as model training and inference. 
\par
In this case, the first pipeline is setup to do data processing, converting the data into long format and splitting it into training and testing partitions, it then trains a single model on the train set $\mathbf z_{1:t-k}$, with $t$ being the total number of observations and $k$ being the length of the forecasting horizon. Finally, the pipeline does a single backtest to calculate the relevant performance metrics. Backtesting refers to the concept of evaluating the performance of a forecasting model based on historical data. Here the model trained on $\mathbf z_{1:t-k}$ does inference for $t-k+1:t$ and is then evaluated on the test set $z_{t-k+1:t}$. 
\newline
The second pipeline is structured similarly in that it does data processing in the same way, leading to an identical data foundation. However, with different data partitions. This is because \texttt{pipeline\_2} was not setup to train a single model, but instead to train multiple models using \ac{HPO} on a train set $\mathbf z_{1:t-2k}$. The \ac{HPO} strategy then does backtests for each trained model on the test set $\mathbf z_{t-2k+1:t-k}$ to determine the best performing one. The resulting, winning hyperparameter configuration is used to train a single model on $\mathbf z_{1:t-k}$, which is evaluated on the validation set $z_{t-k+1:t}$, making it comparable to the model resulting from \texttt{pipeline\_1}.
\par
The \ac{HPO} strategy used in this set of experiments is Bayesian Optimisation as described in \autoref{sec:hpo}. It was not chosen based on any empirical data, which follows in \autoref{sec:expensemble} on the ensemble experiments, but rather due to it being one of the most popular \ac{HPO} strategies, as suggested by \cite{radanliev_review_2023}. This is valid, since the premise of this section is not to provide an optimal strategy, but to show that \ac{HPO} does add a performance improvement to the selected individual algorithm, for which, any \ac{HPO} technique that shows this, suffices. In the tuning workflow of \texttt{pipeline\_2}, Bayesian Optimisation is run with 15 total iterations (\texttt{max\_training\_jobs} $=15$), 5 of which are running in parallel at a time (\texttt{max\_parallel\_jobs} $=5$), leading to only 3 sequential steps. This configuration was not chosen based on empirical evaluations, but rather on internal experience and intuitive heuristics. However, an in depth analysis of optimal configurations is part of \autoref{sec:expensemble}.
\newline
The hyperparameters that are tuned vary by model. For\ac{DeepAR}, \ac{HPO} tunes the learning rate as a continuous hyperparameter parameter and context length as a discrete (integer) parameter. The hyperparameter tuned for \ac{MQ-CNN} is context length. The range of context length is dynamically set based on the given forecast horizon of the time series data used. 
\newline
To ensure robustness, \texttt{pipeline\_1} and \texttt{pipeline\_2} were tested on both \ac{DeepAR} and \ac{MQ-CNN}. Each algorithm was run on all datasets (\texttt{covid\_death}, \texttt{solar}, \texttt{electricity} and \texttt{kaggle\_retail}). The metrics \ac{MAPE}, \ac{MASE}, \ac{WAPE}, \ac{wQL-90}, \ac{wQL-50}, \ac{wQL-10} and \ac{avg-wQL} were computed for each of the resulting $2*2*4=16$ experiments. To simplify the following comparison, the experiments are only compared based on their \ac{avg-wQL}. This choice is based on the fact that every metric used except MASE correlates above 0.98 with the \ac{avg-wQL}, using the standard pearson correlation coefficient, making it a a representative selection, as detailed in \ref{apx:indalgmetriccorr}.
\newline
The comparison is further simplified by computing the mean \ac{avg-wQL} across all datasets for both \ac{MQ-CNN} and \ac{DeepAR}. This leads to a direct comparison between accuracy with and without \ac{HPO} for each algorithm as shown in \autoref{fig:indalgcomp}. The raw results from the experiments can be found in appendix \autoref{apx:indalgrawresults}.

\begin{figure}[H]
    \caption{Individual Algorithm Accuracy With and Without HPO}
    \includegraphics[width=\textwidth]{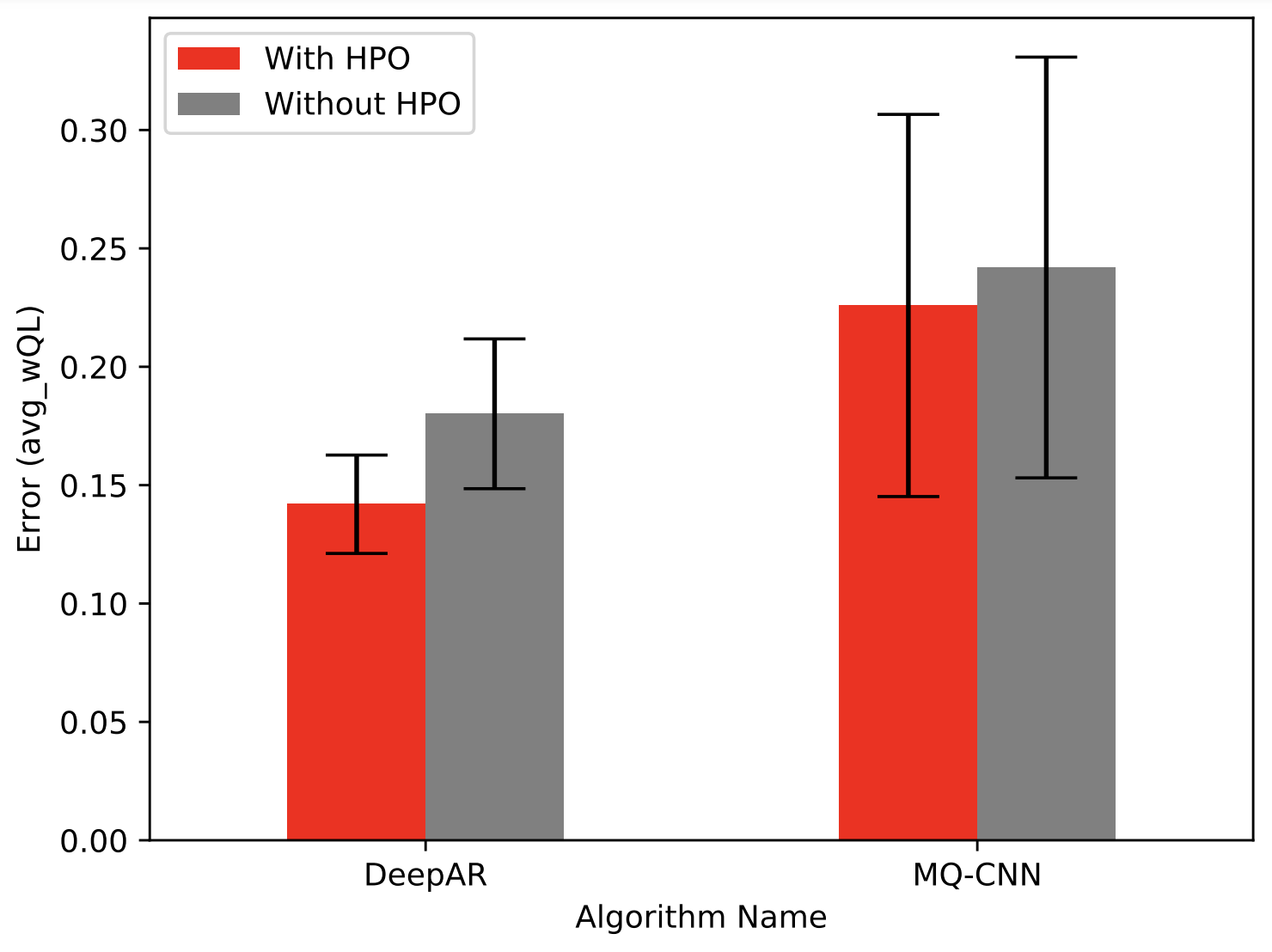}
    \label{fig:indalgcomp}
    \centering
\end{figure}

With 4 observations per experiment, a mean model accuracy with \ac{HPO} of 0.1419 (variance of 0.0208) and a mean accuracy without \ac{HPO} of 0.1801 (variance of 0.0316), adding \ac{HPO} reduces the mean \ac{avg-wQL} for DeepAR by 21.2 \%. For the experiments with and without \ac{HPO} based on MQ-CNN, adding \ac{HPO} leads to a reduction in error (\ac{avg-wQL}) of 6.6 \%.
\newline
These experiments show that \ac{HPO} improves the accuracy for both \ac{DeepAR} and \ac{MQ-CNN}. This shall serve as indication evidence that \ac{HPO} can have a positive impact on algorithm performance, and therefore may also improve the performance of a \ac{AutoML} forecast ensemble as further discussed in \autoref{sec:expensemble}.
\par
Next to the impact of \ac{HPO} on model accuracy, the second interesting aspect is that of training latency. Intuitively the additional models trained as part of hyperparameter tuning are expected to affect the latency and training cost. Even though most literature in the machine learning community only considers accuracy, latency may greatly affect the busyness aspect of an AutoML forecasting platform similar to the one described in \autoref{sec:expensemble}.
\newline
Hence, it makes sense to also consider latency and cost effects of added \ac{HPO}. The former refers to the duration of a workflow (pipeline) or sub-workflow such as the training of a single model (wall time) and the latter refers to the amount of compute resources consumed as part of a workflow (pipeline) or sub-workflow (instance run time). The two may not always correlate due to the varying amount of parallelization utilised in \ac{HPO}. For example, when running two workflows, both using random search, one with \texttt{max\_training\_jobs} $=15$ and \texttt{max\_parallel\_jobs} $=1$ and the second with \texttt{max\_training\_jobs} $=15$ and \texttt{max\_parallel\_jobs} $=15$. They will both have the same \ac{HPO} cost since the number of models trained is 15 in both cases, leading to an identical consumption of resources for \ac{HPO}. The \ac{HPO} latency, however, will differ between the two runs. For the first one with \texttt{max\_parallel\_jobs} $=1$, all models are trained sequentially, leading to the latency being computed as the sum of all individual model training latency. The tuning job with  \texttt{max\_parallel\_jobs} $=15$ will train all models in parallel, leading to the overall latency being computed as the latency of the model with the longest training duration. Assuming all models train for the same duration, this leads to the former being 15 times slower in terms of latency than the latter. 
\newline
Since the experiments outlined in this section all use the same degree of parallelization, latency and training cost can be used interchangeably for comparisons. For better interpretability, the following comparison will be made using latency in seconds. 

\begin{figure}[H]
    \caption{Individual Algorithm Latency With and Without HPO}
    \includegraphics[width=\textwidth]{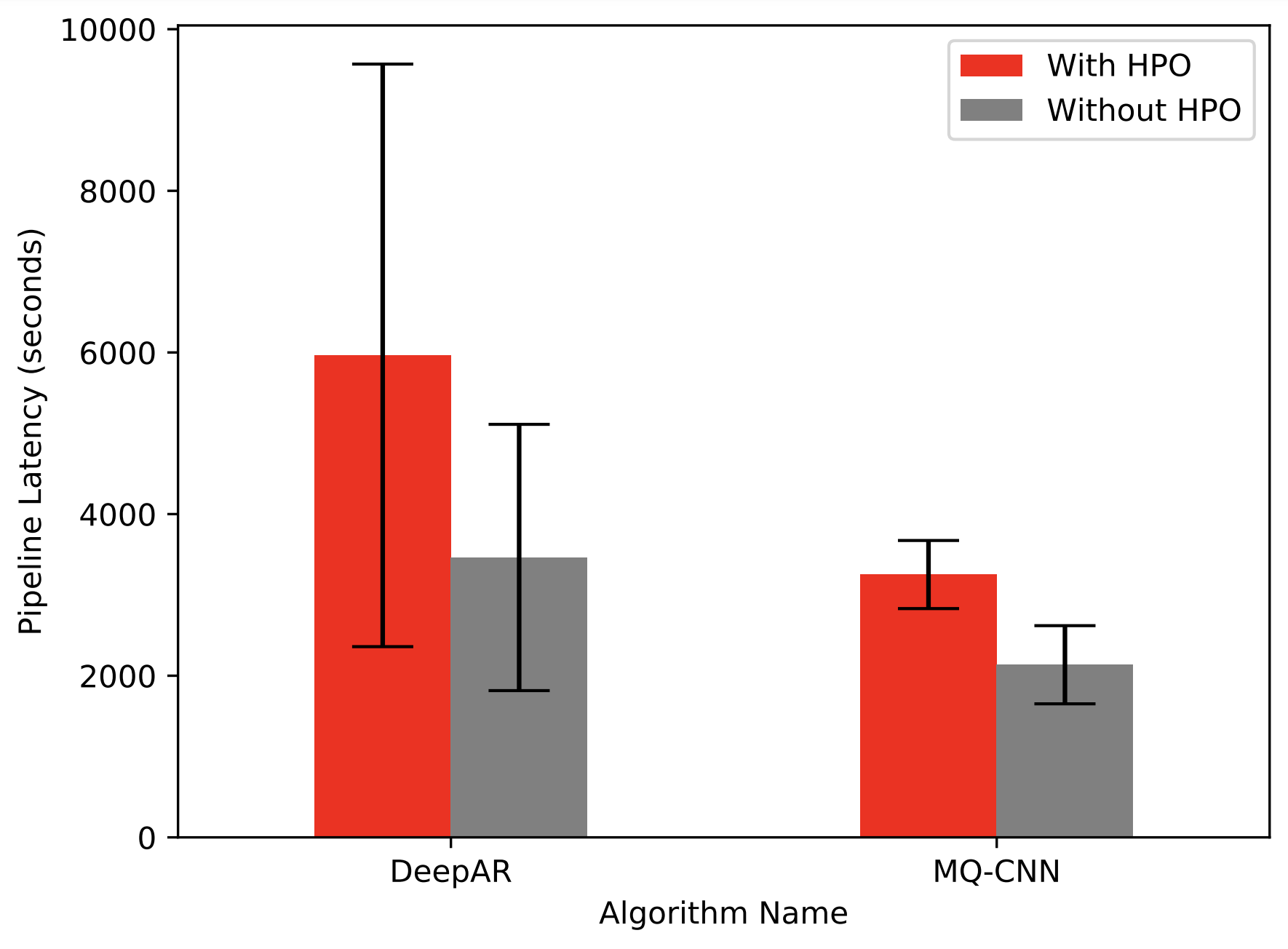}
    \label{fig:indalgcomplat}
    \centering
\end{figure}

The values plotted in \autoref{fig:indalgcomplat} are mean end-to-end pipeline latency with and without \ac{HPO} for both \ac{DeepAR} and \ac{MQ-CNN}. The error bars indicate the standard deviation of latency across datasets. One can recognize the following trends in the plot: Firstly, \ac{DeepAR} is on average 72 \% slower than \ac{MQ-CNN} (with \ac{HPO} it is 83.4 \% slower and without \ac{HPO} it is 62.1 \% slower). Secondly, \ac{HPO} adds an increase of 52.3 \% to the end-to-end pipeline latency, for MQ-CNN, whereas the increase in latency for \ac{DeepAR} when adding \ac{HPO} is 72.2 \%, resulting in an average increase of 62.2 \%. This shows that the accuracy improvement of \ac{HPO} is accompanied by a considerable increase in training latency, indicating the necessity for a trade-off between the two. This is further discussed in following sections.

\subsection{Forecast Ensemble}
\label{sec:expensemble}

This section addresses the core scientific question of this paper by delving into the cost and accuracy effects of various HPO strategies on the performance of a general AutoML forecast ensemble. In particular, this section contains an evaluation of experiments comparing the effect of adding \ac{HPO} to \ac{DeepAR} and \ac{MQ-CNN} on the performance (accuracy and latency) of an ensemble containing \ac{ARIMA}, \ac{ETS}, \ac{NPTS}, Prophet, \ac{MQ-CNN} and \ac{DeepAR}. This paper compares different \ac{HPO} strategies, namely Bayesian and Hyperband (\autoref{exp:hyperbandvsbayesian}) and different variations of these strategies, namely various different numbers of \texttt{max\_training\_jobs}. This parameter refers to the maximum number of models trained as part of \ac{HPO} before the best configuration is determined. The parameter \texttt{max\_parallel\_jobs} refers to the number of models that are trained in parallel as part of hyperparameter tuning. It was not extensively explored and is therefore not the subject of further discussions. In the following experiments \texttt{max\_parallel\_jobs} kept at a constant rate of 5. 
\par
Similarly to the individual algorithm experiments outlined in \autoref{sec:expindalgs}, the forecast ensemble setup was implemented as an end-to-end machine learning pipeline. It consists of 4 steps: (1) data preparation, (2) hyperparameter tuning, (3) ensemble selection (4) final model training. Data preparation and processing are identical to that of \texttt{pipeline\_2} in \autoref{sec:expindalgs}, leading to a training dataset $\mathbf z_{1:t-2k}$, a test dataset $\mathbf z_{t-2k+1:t-k}$ and a validation dataset $\mathbf z_{t-k+1:t}$, which are disjoined. 
\newline
The second step of the pipeline, hyperparameter tuning and model training, is executed in parallel for each algorithm. In the case of the non-deep-learning-based algorithms, that is \ac{ARIMA}, \ac{ETS}, \ac{NPTS} and Prophet, model training is conducted on the training partition using default hyperparameters that have been empirically shown to perform well across datasets, leading to one trained model for each. Depending on the parametrisation of the pipeline, \ac{HPO} may be performed for the deep-learning-based models. If hyperparameters are not to be tuned using \ac{HPO}, then this step follows the same procedure for both \ac{DeepAR} and \ac{MQ-CNN} as for the non-deep-learning-based models. If hyperparameters are to be tuned, the step has the following outline: First, the training dataset $\mathbf z_{1:t-2k}$ with length $l$ is further split into a training partition for tuning $\mathbf z_{1:l-k}$ and a validation partition for tuning $\mathbf z_{l-k+1:l}$. Using a given Hyperparameter tuning strategy, \texttt{max\_training\_jobs} models are trained on the training partition for tuning and evaluated on the validation partition for tuning. The best performing algorithm configuration is then used to train a final model on the entire training dataset $\mathbf z_{1:t-2k}$. This results in one model being trained for each algorithm. 
\newline
In the next step, the ensemble of these algorithms is generated (as outlined in \autoref{sec:ensemble}). Ensemble selection uses the models trained on $\mathbf z_{1:t-2k}$ to make inference on $t-2k+1:t-k$. These predictions, together with the true values in the test partition $\mathbf z_{t-2k+1:t-k}$ are then used to parameterise the ensemble family, resulting in the concrete configuration of the ensemble strategy used for forecasting.  
\newline
The parameters of the ensemble strategy are determined by \ac{HPO}, in particular using the global optimization strategy Basin-Hopping \cite{wales_global_1997}. With the resulting ensembles being evaluated on the validation set $\mathbf z_{t-k+1:t}$. The winning ensemble parameters (local, global and combination parameters) are then used to determine the concrete ensemble to be used based on another set of models (one for each algorithm) trained on training and test sets $\mathbf z_{1:t-k}$ and evaluated on validation. This means, that in this last step, we choose a specific combination of models for each item, resulting in the final ensemble. This ensemble is then evaluated on the validation partition to compute the metrics also used for all the following comparisons in this chapter. The risk of overconfident predictions which is associated with this way of evaluating the ensemble, is recognised, but was not accounted for due to time constraints. 
\par
The previous steps lead to the forecast ensemble pipeline having the following parameters and input variables: the dataset used for training, a binary which determines if hyperparameters are to be tuned or if default ones should be used, and in the former case, additional parameters for the maximum number of models trained by \ac{HPO}, the maximum number of models trained in parallel and the \ac{HPO} strategy. After initial exploration, all other parameters where regarded as static, logically set, or are learned by the pipeline and shall not be a further subject of this paper.
\par
All experiments, testing a particular configuration of pipeline parameters were repeated with three different random seeds for each of the four datasets, yielding a total of 12 runs per experiment, to ensure more rigorous results.

\subsection{Impact of Default HPO on the Ensemble}
\label{exp:compindensemble}

\begin{figure}[htbp]
    \caption{Ensemble Performance With and Without HPO}
    \begin{subfigure}{.5\textwidth}
        \centering
        \includegraphics[width=.9\linewidth]{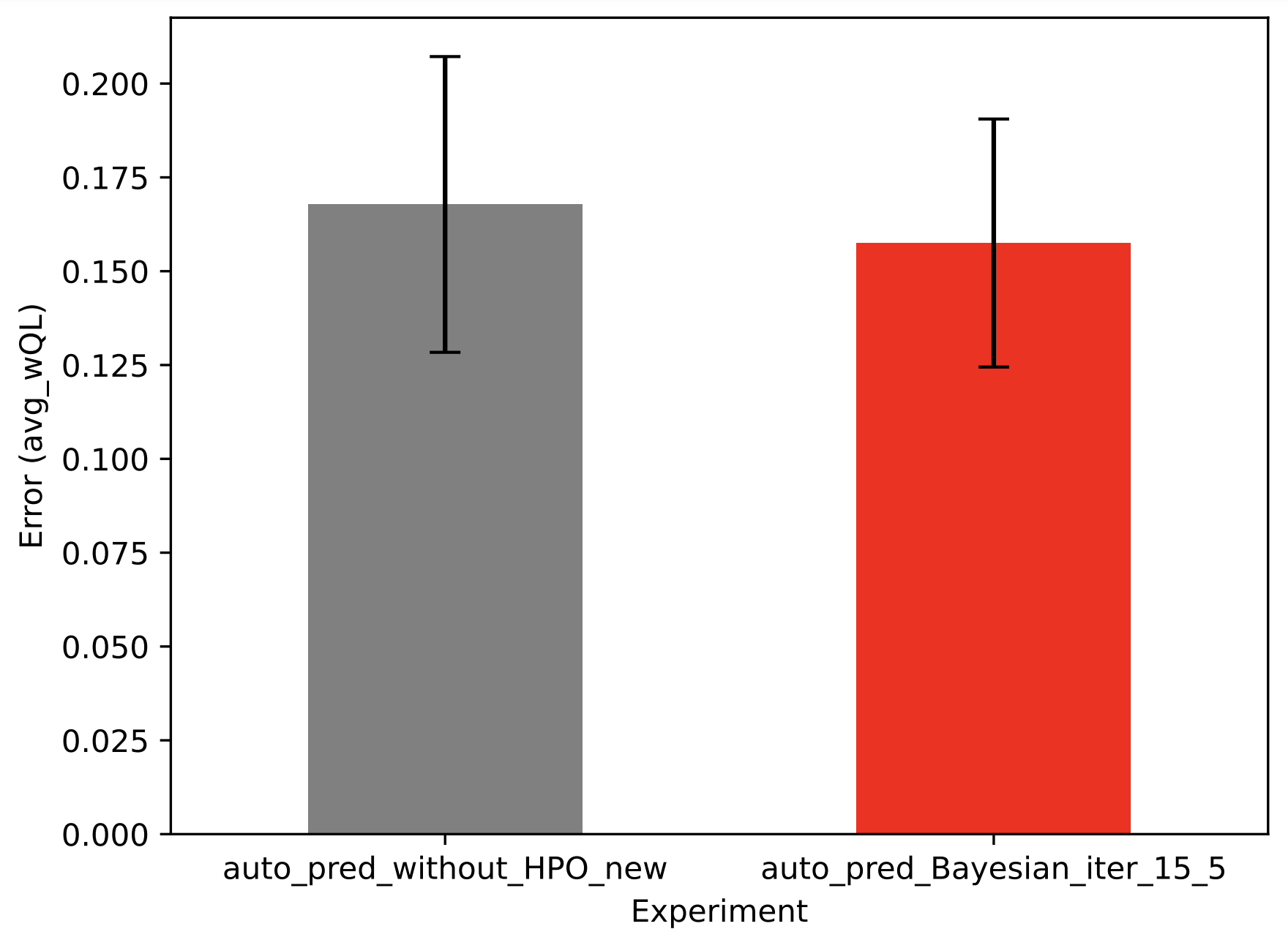}
        \caption{Accuracy}
        \label{fig:ensemblewwohpo:accuracy}
    \end{subfigure}%
    \begin{subfigure}{.5\textwidth}
        \centering
        \includegraphics[width=.9\linewidth]{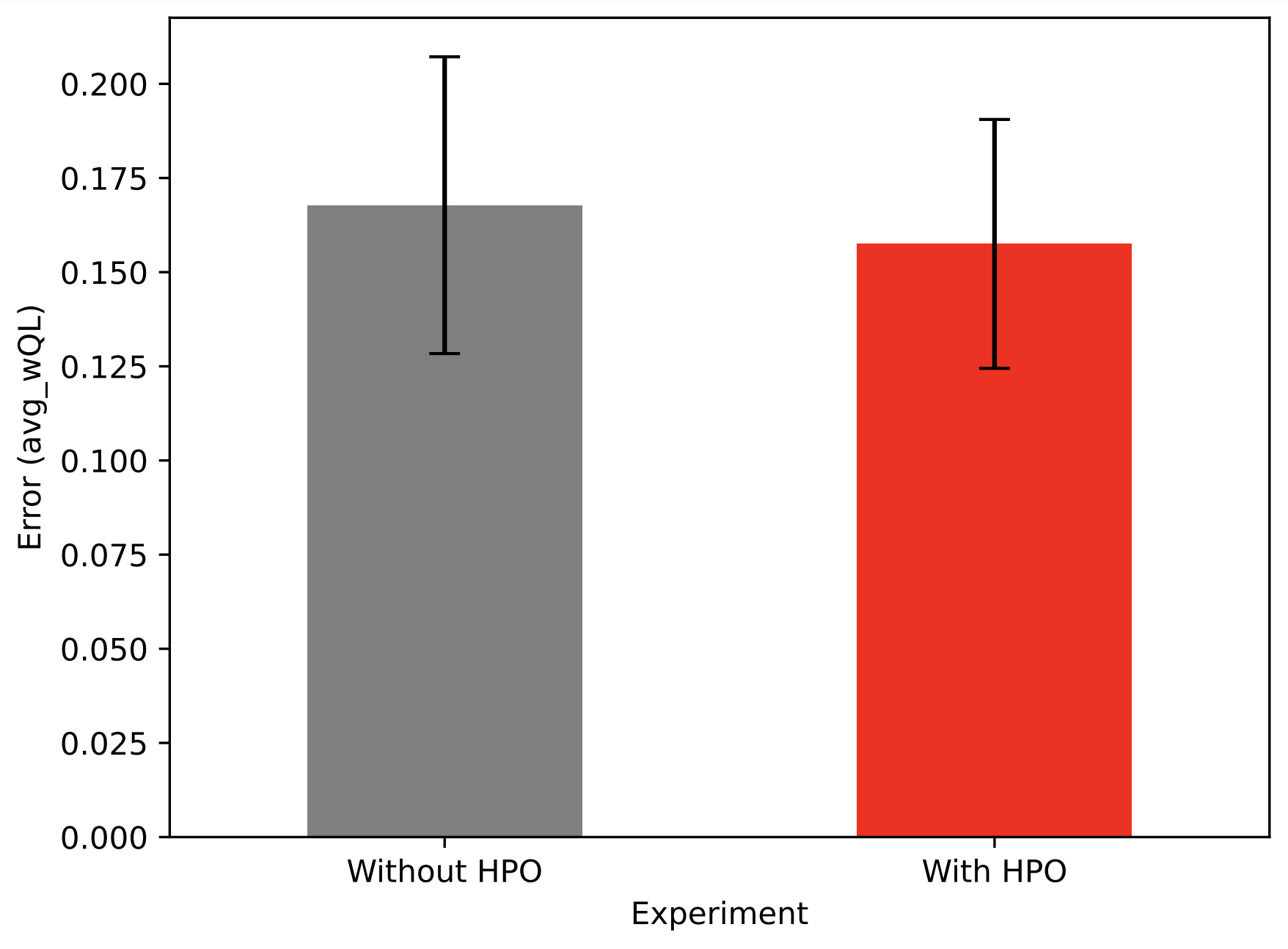}
        \caption{Pipeline Latency}
        \label{fig:ensemblewwohpo:latency}
    \end{subfigure}
    \label{fig:ensemblewwohpo}
    \centering
\end{figure}

Before delving into the empirical evaluation of different \ac{HPO} strategies and their configurations in the following subsections, this one will look at a comparison between the effects of \ac{HPO} on individual algorithms as outlined in \autoref{sec:expindalgs} and its effects on the final forecast ensemble. 
\newline
To make the magnitude of the effect on the ensemble comparable to the effect on individual algorithms, this subsection is going to look at experiments conducted with the \ac{HPO} strategy being Bayesian, the number of models trained for tuning \texttt{max\_training\_jobs} $=15$ and \texttt{max\_parallel\_jobs} $=5$, as was done in the experiment including individual algorithms outlined in \autoref{sec:expindalgs}.
When considering the effect of adding \ac{HPO} to \ac{DeepAR} and \ac{MQ-CNN} in the ensemble, as shown in \autoref{fig:ensemblewwohpo}, it is apparent that differences are more marginal compared to those observed for the individual algorithms.  

\begin{figure}[htbp]
    \caption{Individual Algorithm vs. Ensemble}
    \includegraphics[width=.7\textwidth]{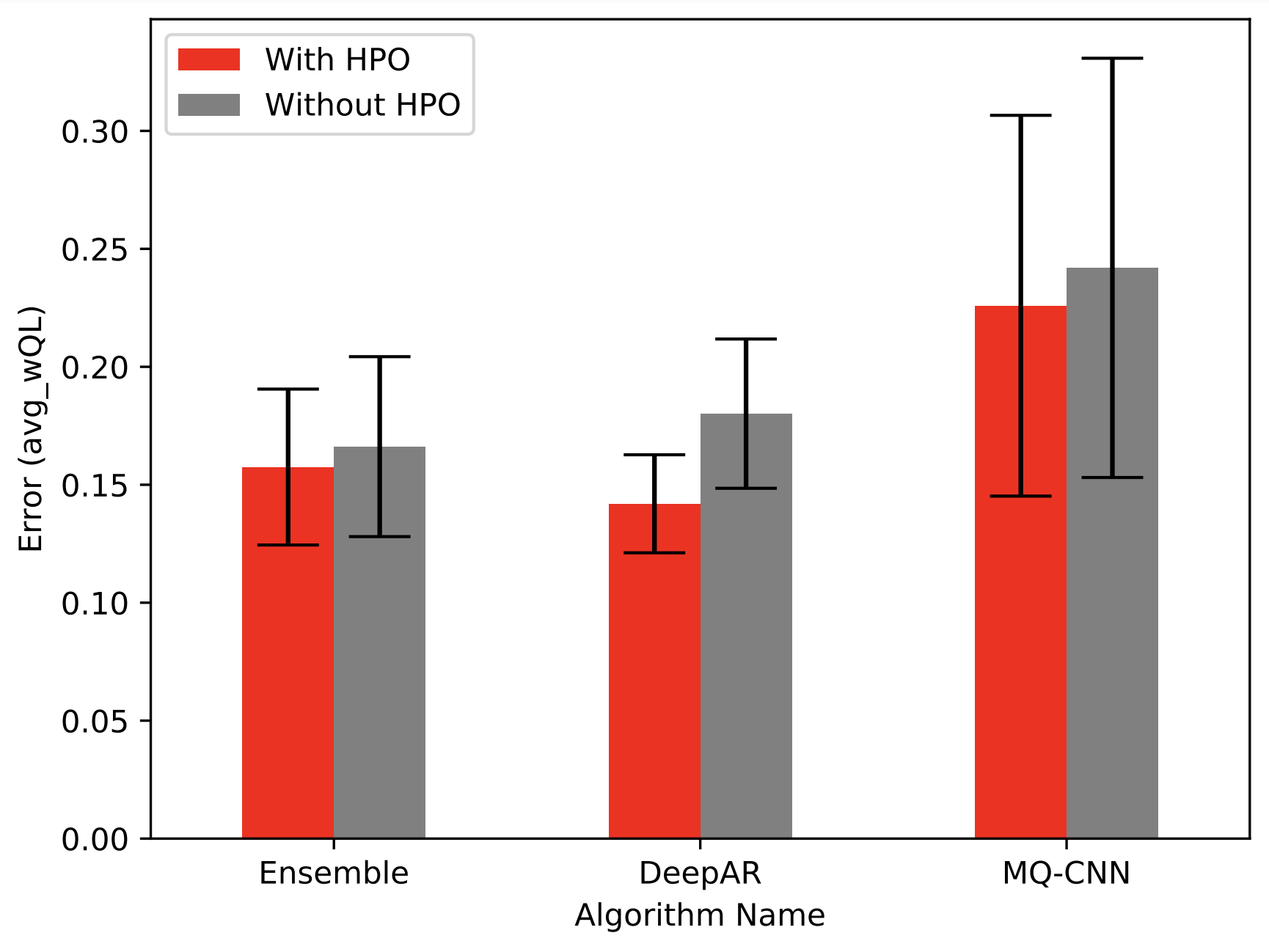}
    \label{fig:indensemblecomp}
    \centering
\end{figure}

This is both true for the difference in error, with \ac{HPO} leading to a 6.1 \% decrease, visible in \autoref{fig:ensemblewwohpo:accuracy} and the difference in end-to-end pipeline latency, with a 14.0 \% increase caused by adding \ac{HPO}. The former can be partly explained by the added improvement of other algorithms included in the ensemble (\ac{ARIMA}, \ac{ETS}, \ac{NPTS}, Prophet). For example, if results from these other algorithms improve the forecast, then this diminishes the positive accuracy effect on the final forecast of adding \ac{HPO} to \ac{DeepAR} and \ac{MQ-CNN}. The latter difference can be explained by overall higher latency of the pipeline compared to the one used for the individual algorithms, caused by extra steps such as ensembling. These extra steps increase the overall pipeline latency and therefore reduce the relative increase in latency caused by \ac{HPO}. 
\newline
The differences in effect size of adding \ac{HPO} to the deep-learning-based algorithms is visualized in \autoref{fig:indensemblecomp}, which compares the resulting accuracy change for individual algorithms as well as the ensemble.

This comparison shows that the ensemble pipeline without \ac{HPO} leads to 19.5 \% more accurate results on average, than both \ac{DeepAR} (8.4 \% higher error) and \ac{MQ-CNN} (45.6 \% higher error) individually without \ac{HPO} (grey bars), demonstrating the positive effect of ensemble learning on forecast accuracy. When comparing the effect size of adding \ac{HPO} to the individual algorithms and the ensemble, it is noticeable that this effect is smaller for the ensemble due to the aforementioned reasons. To be precise, for \ac{DeepAR}, \ac{HPO} improves accuracy by 21.2 \% and for \ac{MQ-CNN} 6.6 \%, leading to an average improvement for the individual deep-learning based algorithms of 13.9 \%. Whereas the accuracy improvement of adding \ac{HPO} to the deep-learning algorithms on the final ensemble is only 5.2 \% using Bayesian Optimisation with 15 models trained for tuning, five of which in parallel. It is further noteworthy that \ac{DeepAR} with \ac{HPO} has the overall lowest accuracy in this comparison. The reason for this is not clear. However, it would make sense to assume that the trade-off between rigorous results and high test performance in ensemble strategy selection, could be the cause. 
\newline
These results indicate that using the ensemble pipeline provides better results than the individual algorithms without \ac{HPO} and further that adding \ac{HPO} to the ensemble pipeline can further improve its performance.

\subsection{Experiment Configuration Overview}
\label{exp:experiment overview}

\begin{figure}[H]
    \caption{Experiment Overview}
    \includegraphics[width=.8\textwidth]{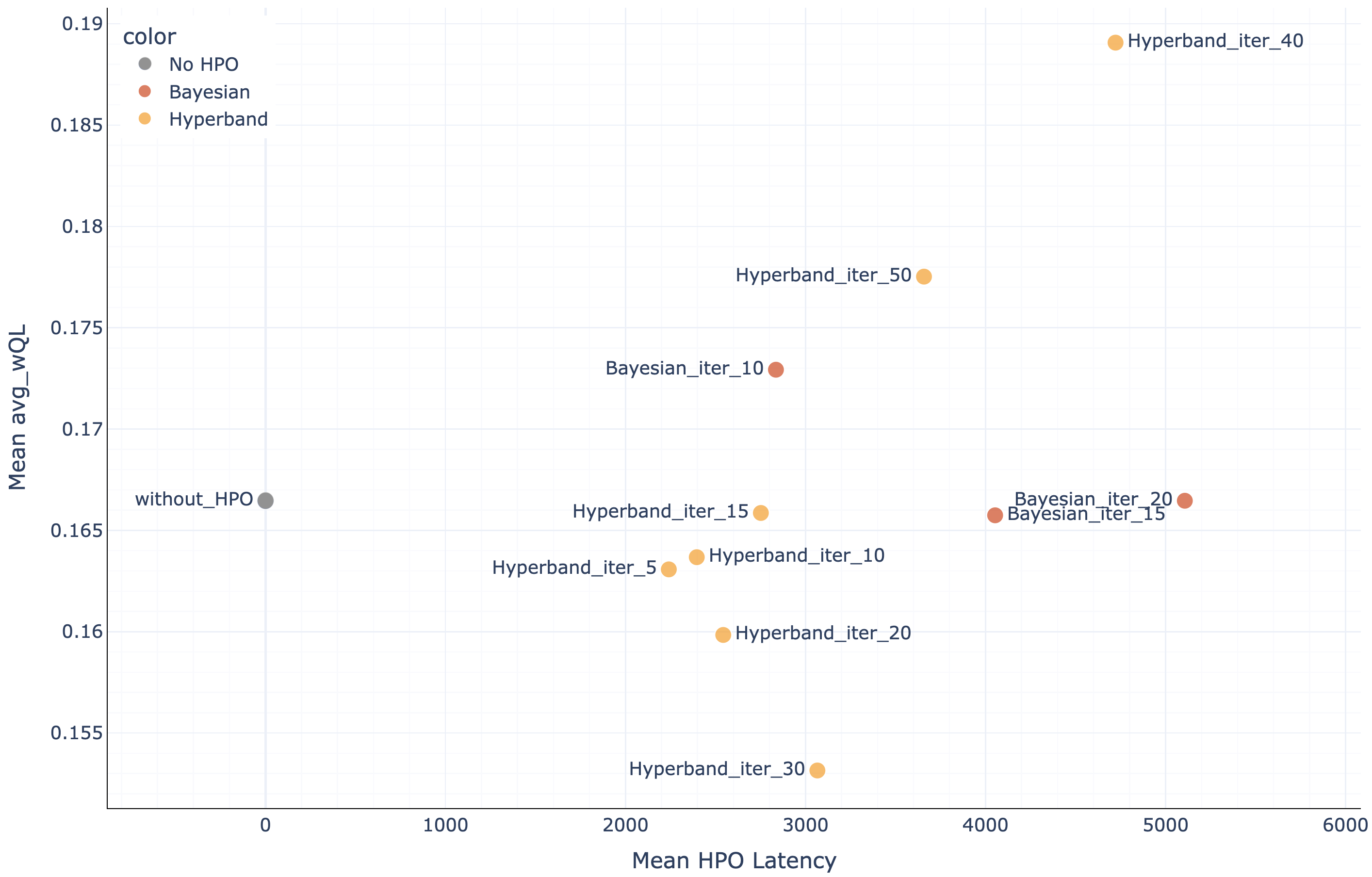}
    \label{fig:expoverview}
    \centering
\end{figure}

To show that the positive impact on the ensemble performance of adding \ac{HPO} to the deep-learning-based algorithms can be further increased, other optimisation strategies as well as configurations of those where tested. The following subsection provides an overview of the conducted experiments. This includes experiments with Bayesian Optimisation and Hyperband as well as permutation of these strategies with different numbers of \texttt{max\_training\_jobs}. As mentioned before, each concrete configuration is repeated 12 times (on 4 different datasets, with 3 different random seeds each).
\newline
An overview of all experiments conducted can be seen in \autoref{fig:expoverview}, with the x-axis being the mean latency \ac{HPO} of a given configuration and the y-axis being the mean \ac{avg-wQL}. This figure is to be read such that one is ideally looking for a configuration on the bottom left with low latency and low \ac{avg-wQL}. Throughout this paper, configurations without \ac{HPO} are grey, those with Hyperband are orange and the Bayesian Optimisation ones red. Note that the figure is not comparing the overall pipeline latency here, but rather the latency of hyperparameter tuning, since the latency of components other than \ac{HPO} is assumed to stay relatively constant. 
\newline
The evaluation of results is detailed in the following sections. Looking only at \ac{HPO} makes the effects of different configurations more pronounced and therefore easier to evaluate. To make the direct comparison of different numbers of \texttt{max\_training\_jobs} easier, the number of \texttt{max\_parallel\_jobs} is kept at a constant of 5. A table with the results of all experiments can be found in \autoref{apx:enemblerasresults}.

As the experiments are mainly differentiated by two characteristics, namely tuning strategy and number of training jobs \texttt{max\_training\_jobs}, it makes sense to analyse the results along these two dimensions. This is advantageous as it breaks the problem of determining an optimal configuration down into two simpler problems: (1) find an optimal strategy and (2) find an optimal setting of \texttt{max\_training\_jobs} for that strategy. In an ideal case, optimal refers to a strategy that is both faster and more accurate than any other reasonable configuration. Such an optimum could not be found as part of this paper and is unlikely to exist because model accuracy positively depends on training time for low latencies (ambivalent for higher latencies as they may increase model variance). This paper instead aims at finding a reasonable trade-off between accuracy and latency, as defined in \autoref{exp:strategyvariations}. 
A comparison between Hyperband and Bayesian optimisation is conducted in the following \autoref{exp:hyperbandvsbayesian} and an extension of this comparison, by varying the number of \texttt{max\_training\_jobs}, is described in \autoref{exp:strategyvariations}. 

\subsection{Hyperband vs. Bayesian Optimisation}
\label{exp:hyperbandvsbayesian}

This subsection compares the performance of the ensemble pipeline with \ac{HPO}, either using Hyperband or Bayesian Optimisation as a tuning strategy. There are several approaches in which one can execute this comparison, such as comparing a single sample configuration, as was done in the individual algorithm evaluation in \autoref{sec:expindalgs} and in \autoref{exp:compindensemble} on the impact of default \ac{HPO} on the ensemble. This approach may have been valid there, as these comparisons have the sole purpose of showing that an effect exists for which any configuration that does this suffices. This section, however, aims at answering the question of a generally better strategy in the context of the \ac{AutoML} forecast ensemble, requiring a more representative approach. This is achieved by comparing the two strategies for multiple configurations respectively. 

\begin{figure}[H]
    \caption{Experiment Overview Accuracy}
    \includegraphics[width=.9\textwidth]{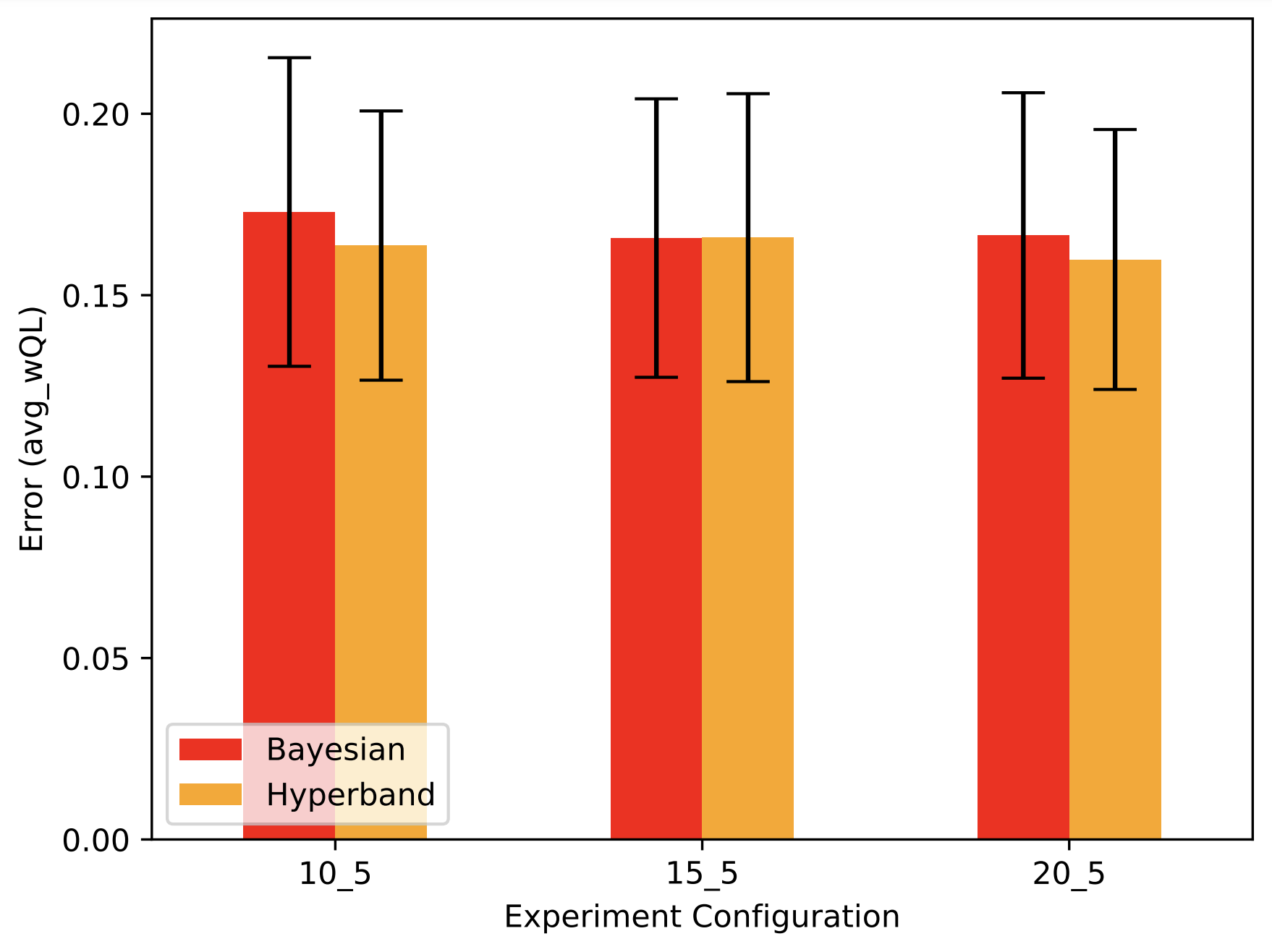}
    \label{fig:bayesianvshyperband}
    \centering
\end{figure}

The error comparison of the final ensemble with the two strategies couldn't show that there is a substantial difference between Hyperband and Bayesian Optimization. This was determined by comparing both strategies for 3 different configurations of \texttt{max\_training\_jobs} and \texttt{max\_parallel\_jobs}. To be precise, Hyperband 10;5 is compared to Bayesian 10;5, Hyperband 15;5 is compared to Bayesian 15;5 and Hyperband 20;5 is compared to Bayesian 20;5, with the first number always referring to the number of \texttt{max\_training\_jobs} and the second number denoting the number of \texttt{max\_parallel\_jobs}. This comparison is shown in \autoref{fig:bayesianvshyperband}.
\par
The difference between Hyperband and Bayesian is minor for all variations of \break \texttt{max\_training\_jobs} and \texttt{max\_parallel\_jobs} tested (10;5, 15;5 and 20;5). This does not mean that \ac{HPO} with Bayesian Optimisation has the same effect on the accuracy of the individually tuned algorithms as \ac{HPO} with Hyperband, but rather indicates that no large effect on the accuracy of the final ensemble can be detected. This may be due to the relatively small number of 4 repetitions with different random seeds for each experiment. Hence, no optimization strategy can be regarded as much better based on its accuracy. One could, however, interpret the fact that the mean error of Hyperband is on average 3.1 \% lower than that resulting from Bayesian Optimisation, as an indicator that the former may be better.
\par
\begin{figure}[H]
    \caption{Experiment Overview Latency}
    \includegraphics[width=.9\textwidth]{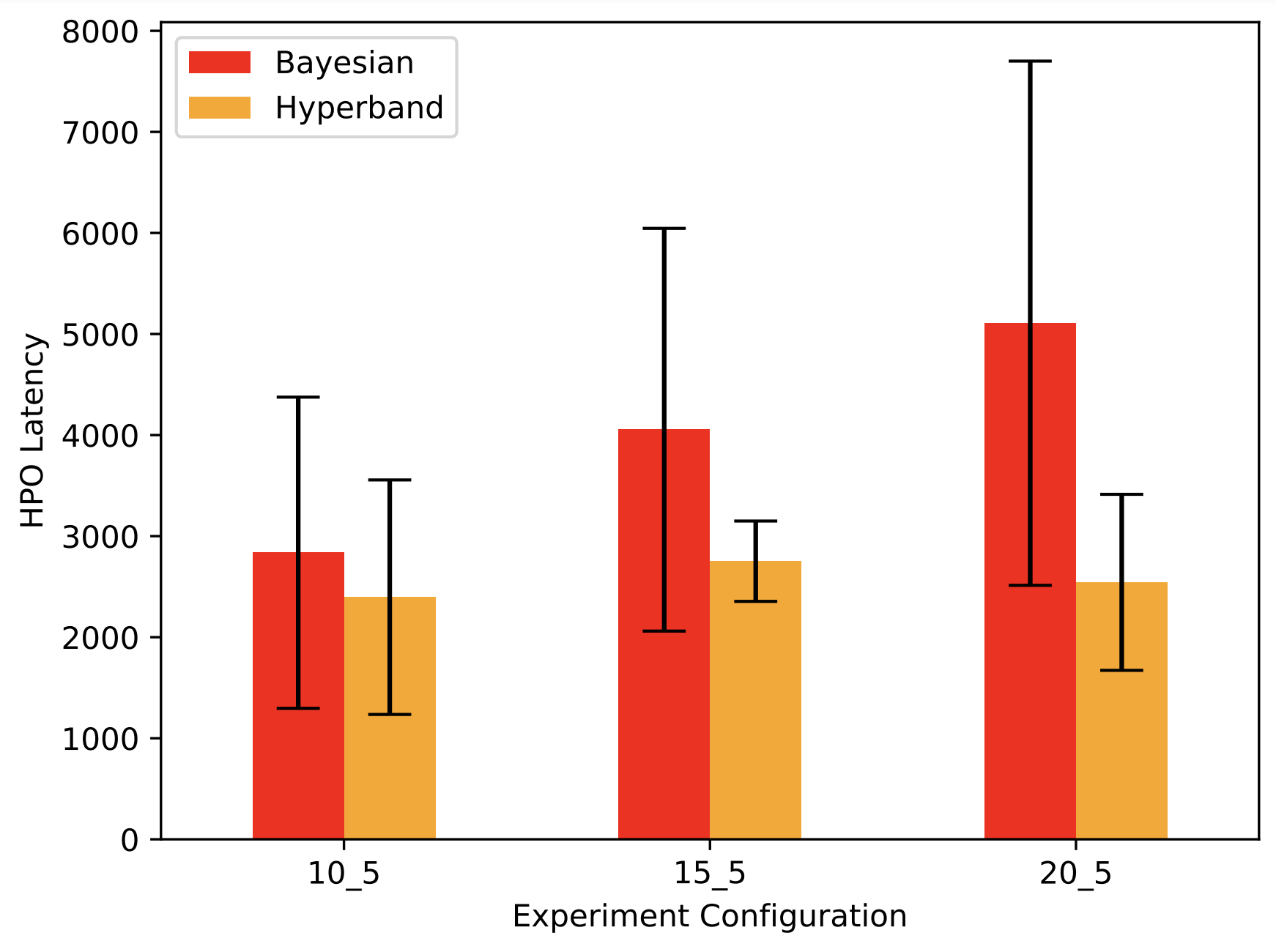}
    \label{fig:bayesianvshyperbandlatency}
    \centering
\end{figure}

When comparing the ensemble performance based on latency, Hyperband outperforms Bayesian Optimisation. The design of the latency comparison is structured the same as the accuracy comparison, with, Bayesian  \texttt{max\_training\_jobs}$_i$; \texttt{max\_parallel\_jobs}$_i$ being compared to Hyperband \texttt{max\_training\_jobs}$_i$; \texttt{max\_parallel\_jobs}$_i$ as shown in \autoref{fig:bayesianvshyperbandlatency}.

The comparison shows that for Hyperband, the mean latency of Hyperband is consistently lower than that of Bayesian Optimisation. It is further noticeable that Hyperband latency doesn't increase much with higher numbers of models trained for tuning. For 15 trained models, the \ac{HPO} latency of Hyperband is only 14.9 \% higher than the latency for 10 models. Going from 15 to 20, the latency even decreases again by 7.6 \% (which is expected to be due to random variation and not to be an effect of the increased number of models), overall, leading to a standard deviation between latencies of different configurations of 146.1 seconds, whereas the standard deviation latencies for Bayesian optimisation is higher with 928.0 seconds. Here we see a clear increase in latency with higher numbers of trained models for tuning (42.9 \% from 10 to 15 trained models and another 26.0 \% from 15 to 20 trained models). This makes Hyperband favorable over Bayesian optimisation in terms of latency as it is on average 32.6 \% faster.
\newline
Overall, this makes Hyperband more suitable for use in the \ac{AutoML} forecast ensemble, indicating both slightly better accuracy results as well as a lower latency. This is the reason why the focus of further experiments that are part of \autoref{exp:strategyvariations} is on Hyperband.

\subsection{Tuning Strategy Variations}
\label{exp:strategyvariations}

This section will explore \ac{AutoML} forecast ensemble pipeline results with different variations of resource allocations for \ac{HPO}, focusing on the Hyperband optimisation strategy. As suggested by \cite{li_system_2020}, Hyperband can be parallelized in a way that doesn't change training cost but reduces latency (if \texttt{max\_training\_jobs} remains constant at 5 and only \texttt{max\_parallel\_jobs} is increased). Such variations of \texttt{max\_parallel\_jobs} are not assumed to affect the accuracy of the final ensemble. Therefore, this analysis will only consider the total number of jobs, for the sake of simplicity. Trade-offs between latency and cost (meaning \texttt{max\_training\_jobs} and \texttt{max\_parallel\_jobs}) can be arbitrarily chosen for particular usecases.  

\begin{figure}[H]
    \caption{Hyperband Metric Overview}
    \includegraphics[width=.9\textwidth]{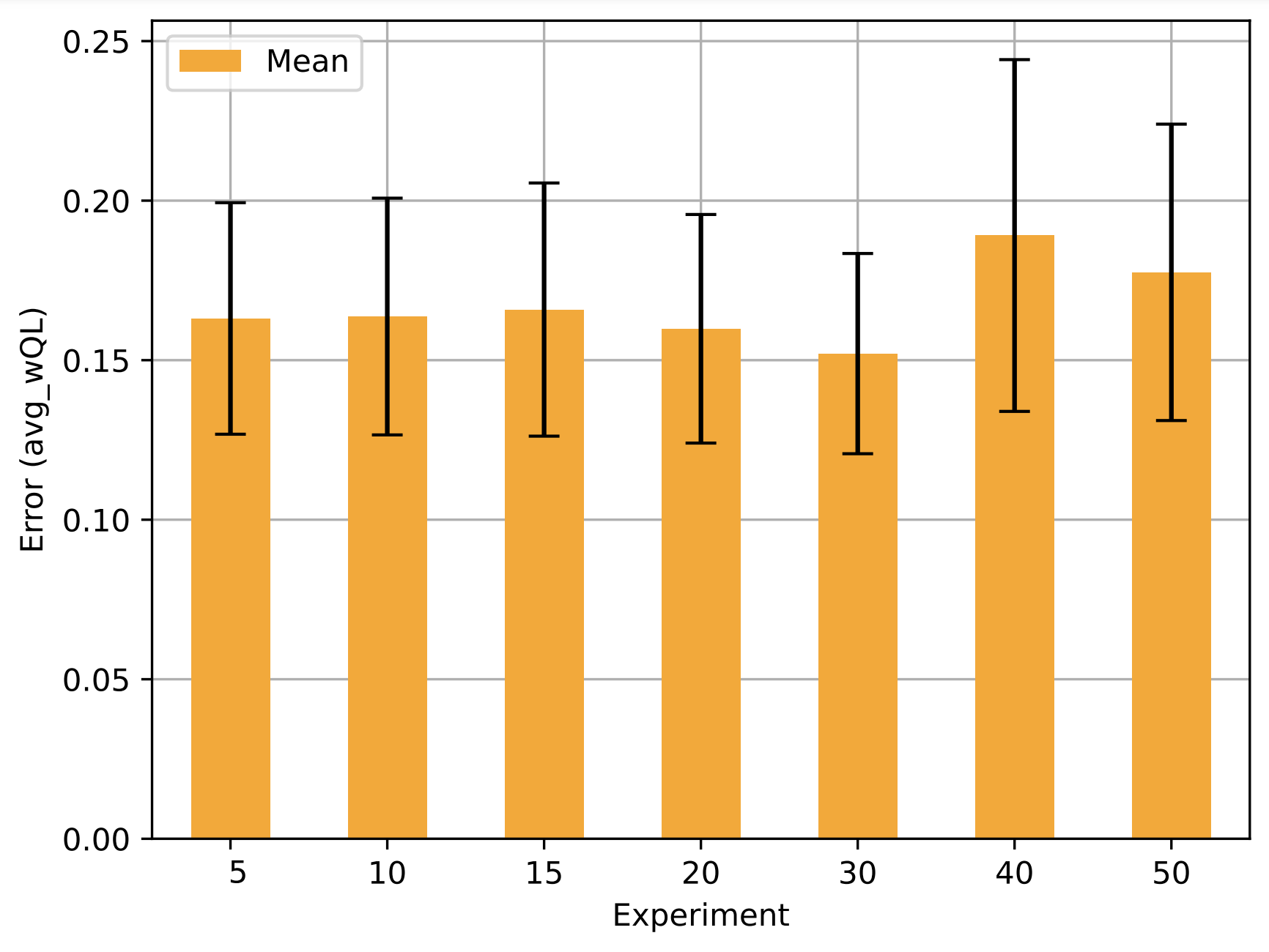}
    \label{fig:hyperbandmetriccomp}
    \centering
\end{figure}

The variations in accuracy resulting from different numbers of \texttt{max\_training\_jobs}, are shown in \autoref{fig:hyperbandmetriccomp}. These differences are measured as differences in the mean of the distributions of all experiments performed with a particular configuration (in the plot this distribution is indicated by its mean and standard deviation). The bars are each labeled with two numbers separated by an underscore. The x-axis refers to the number \texttt{max\_training\_jobs}. The number of \texttt{max\_parallel\_jobs} is kept at a constant rate of 5. 
\par
In the above plot one can see that for \texttt{max\_training\_jobs} $\in \{5,10,15\}$ the \ac{avg-wQL} stays relatively constant. For a higher number of iterations, e.g. \texttt{max\_training\_jobs} $\in \{20,30\}$, it then decreases, improving model performance, until then increasing again for \texttt{max\_training\_jobs} $\in \{40,50\}$, potentially due to hyperparameter over-fitting. Meaning that the hyperparameters selected perform very well during training, but don't generalize well to the validation partition for tuning, leading to a higher error. 
The configuration with the lowest mean error across all experiments and repetitions is that with 30 trained models. It is 4.9 \% better than the second best configuration with 20 models trained for tuning and 19.6 \% better than the worst configuration test with 40 models trained. 
\par
\begin{figure}[H]
    \caption{Hyperband Latency Overview}
    \begin{subfigure}{.5\textwidth}
        \centering
        \includegraphics[width=\linewidth]{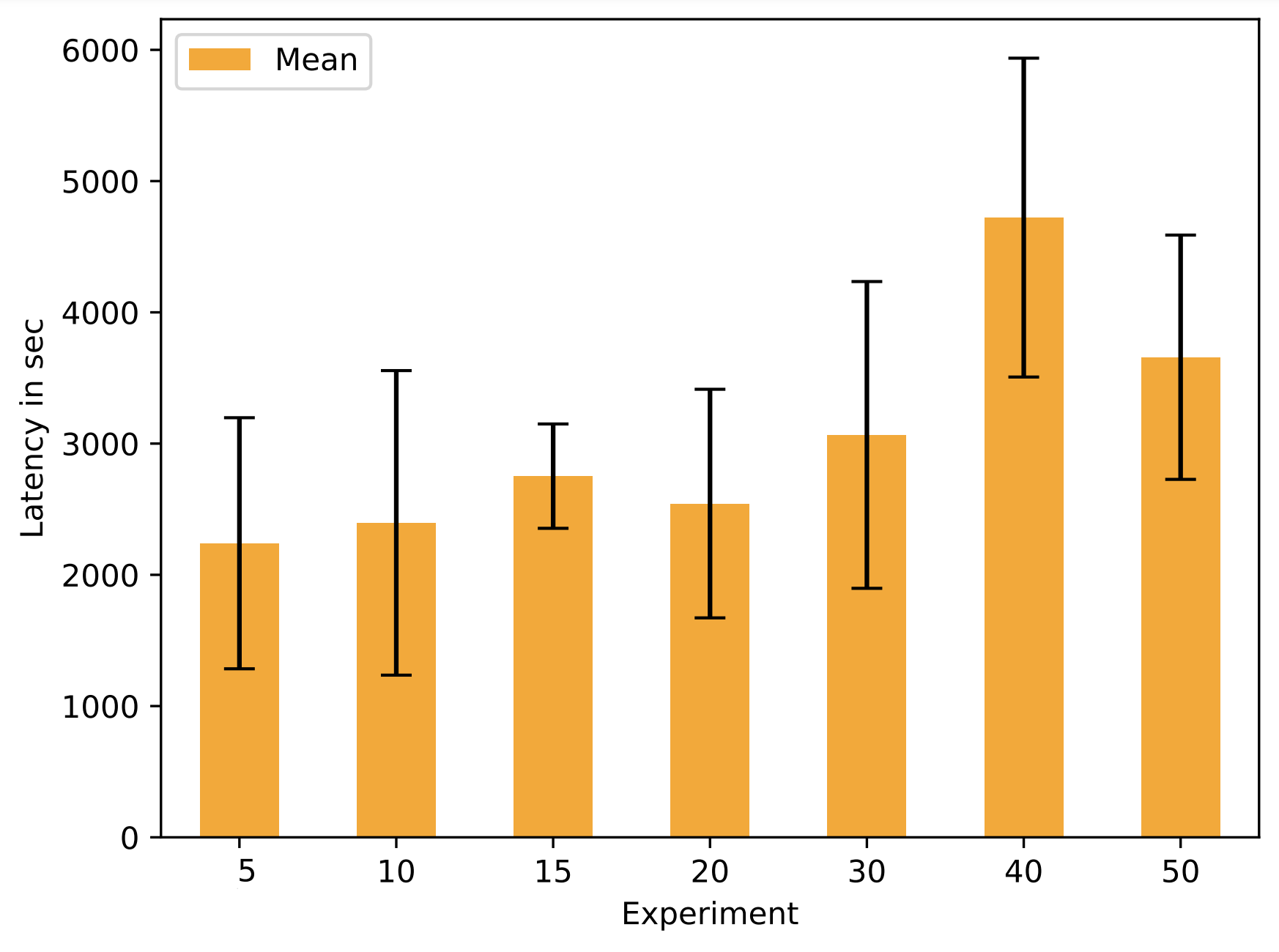}
        \caption{Average Across Datasets}
        \label{fig:hyperbandavglatency}
    \end{subfigure}%
    \begin{subfigure}{.5\textwidth}
        \centering
        \includegraphics[width=\linewidth]{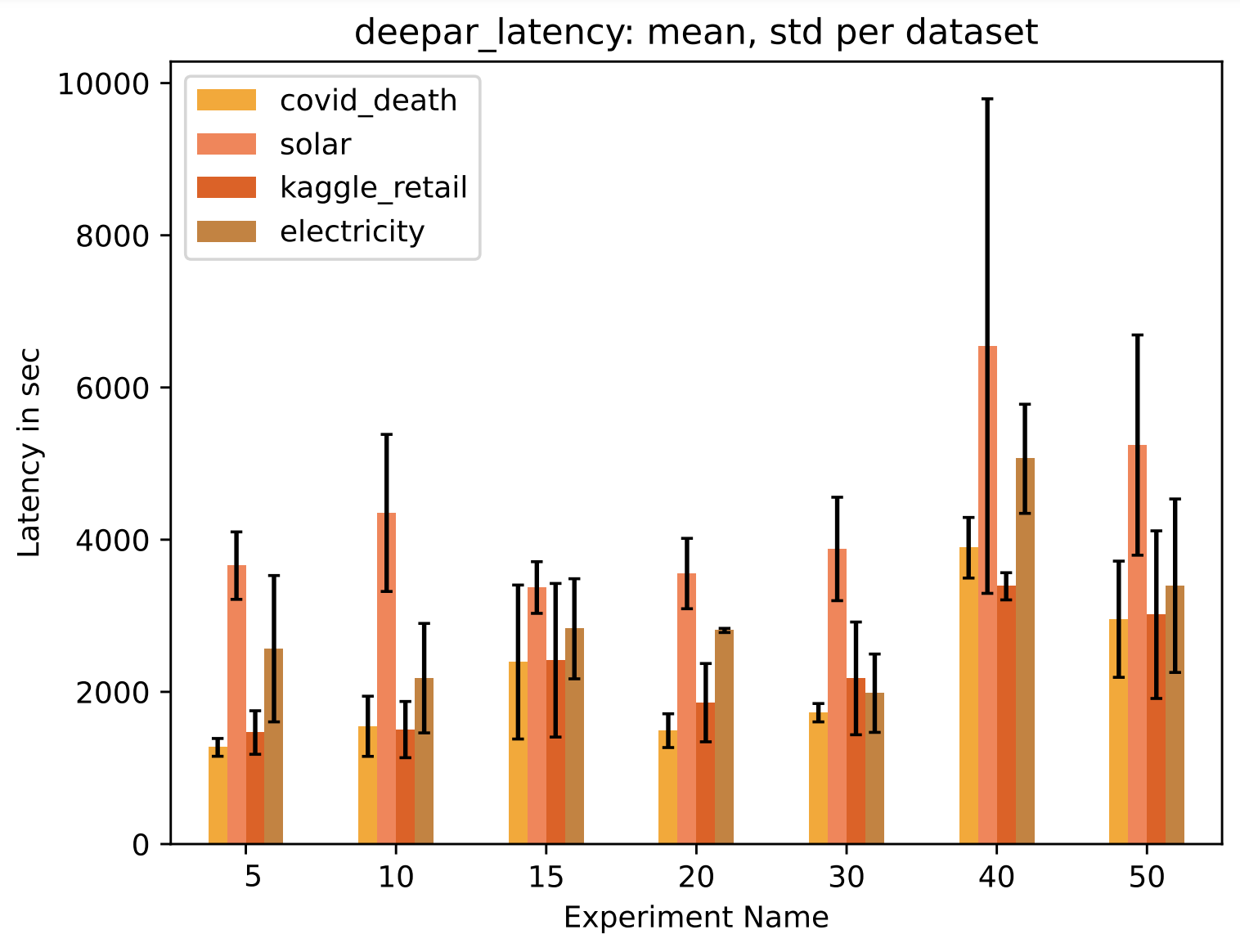}
        \caption{Per Dataset Latency}
        \label{fig:hyperbandperdatasetlatency}
    \end{subfigure}
    \label{fig:hyperbandlatency}
    \centering
\end{figure}

When considering the latency differences of the Hyperband configurations, one can observe an increase with higher numbers of trained models. This behaviour is expected, as with a constant number of \texttt{max\_parallel\_jobs} an increased number of  \texttt{max\_training\_jobs} leads to more sequential training jobs, which increases latency. This trend is not monotonic, which could be partly attributed to the randomness in the dynamic early stopping behaviour of Hyperband. An example of this would be that a tuning job which is supposed to try 20 different configurations, stops 15 of them very early, because they are not promising. A tuning job with 10 configurations may only terminate 5 early, because they are not promising. This leaves both variations with 5 models left to train, minimizing the latency difference between them, or possibly even reverting it. 
\newline
Figure \ref{fig:hyperbandavglatency} shows the mean latency across datasets for different numbers of models trained per tuning job, with the error bars indicating the standard deviation (of the different means across runs for each dataset, with variation within repetitions of a single dataset being disregarded). This loss of information is remedied in \autoref{fig:hyperbandperdatasetlatency}, which shows a more granular latency comparison at a per dataset level.

The latency comparison on the dataset level, also reflects the qualities of the different datasets, namely number of items and number of time steps. The solar dataset monotonically has the highest mean latency for all of the evaluated configurations, which can be presumably attributed to it also having the highest number of time steps (that is, 7033). This hints at the fact that the number of time steps is more indicative of tuning (and therefore also training) latency than number of items, or number of data points ($n\_items*n\_time\_steps$), since solar is not first in either of those. This is supported by the fact that the electricity dataset, which has the highest number of data points (1491840, which is 54.8\% more than the solar dataset with 963521) and the second highest number of time steps (4032), has the second highest latency for all configurations except one. This difference in tuning latency is not as pronounced for the smaller datasets, covid death (with 48760 time steps) and kaggle retail (with 90936 time steps), as these two are comparable in latency for most datasets, with one or the other being faster at times. 
\par
As can be expected, the smallest number of total models trained during tuning also has the smallest latency. As the optimal configuration in terms of latency (5;5) does not coincide with the optimal configuration for accuracy (30;5). This means a trade-off has to be made. This trade of can be conceptualised as a discrete linear optimisation problem \autoref{for:optimalconfig}, where $latency(\cdot)$ and $error(\cdot)$ return the normalised latency and error respectively and  $\theta_0 \in [0,1]$ being a weight that determines how much latency and error should each affect the outcome. 

\begin{equation}
    \min (\theta_0*latency(conf_i)+(1-\theta_0)*error(conf_i))
    \label{for:optimalconfig}
\end{equation}

These weights should be set depending on the business use case at hand. For example, if latency is not important at all, $\theta_0$ is set to $0$, leading to Hyperband with \texttt{max\_training\_jobs} $=30$. If one primarily cares about latency and error is only of secondary importance, then one can set $\theta_0=1$, leading to Hyperband with \texttt{max\_training\_jobs} $=5$ being optimal. It is important to note, that even though latency and error are normalized, $\theta$ is not necessarily a linear indicator. Meaning that a $\theta$ does not directly mean we equally care about latency and error, since a non-symmetric distribution in either may skew results.
\par
\begin{figure}[H]
    \caption{Accuracy vs. Latency Trade-Off}
    \includegraphics[width=.97\textwidth]{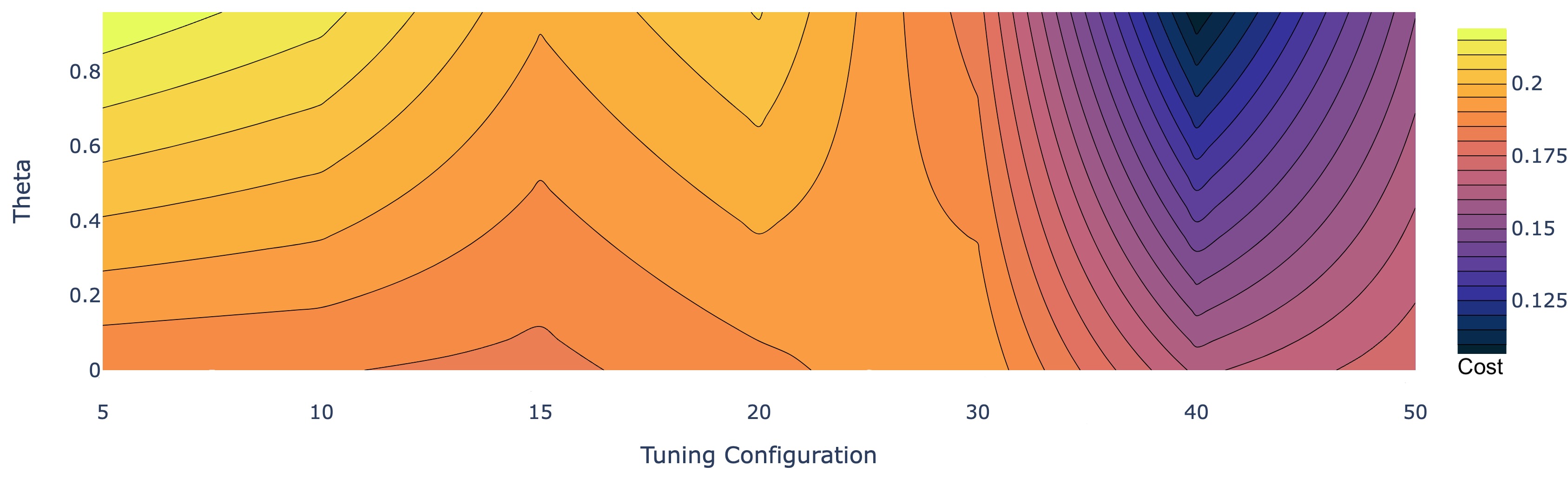}
    \label{fig:thetatradeoff}
    \centering
\end{figure}

An analysis of \autoref{for:optimalconfig} shows that for $\theta_0 \le 0.1798$ the configuration Hyperband with \texttt{max\_training\_jobs} being 30 is optimal, for $\theta_0 > 0.1798$ the configuration Hyperband with \texttt{max\_training\_jobs} being 5 is optimal. This relationship is visualized in \autoref{fig:thetatradeoff}, which shows the value of \autoref{for:optimalconfig} for the continuous variable $theta$ (y-axis) and the discrete variable $conf_i \in \{5,10,15,20,30,40,50\}$ (x-axis), which indicates the \texttt{max\_training\_jobs}. The color associated with a given tuple of tuning configuration and theta shows the cost of \autoref{for:optimalconfig}, with lighter colors indicating a lower cost (better configuration).

The \hyperref[fig:thetatradeoff]{figure} is to be read horizontally with the optimal configuration being that where the cost is lowest (where the color is lightest). In the figure, the number of \texttt{max\_training\_jobs} $=30$ is best for all $\theta$ below $~2.8$. This is due to the low frequency of the contour lines, which was chosen to increase readability. With a higher frequency, one could better approximate the true cutoff point visually (this was not done because it makes it harder to differentiate the colors used and therefore also to read the figure). The plot instead has the purpose of roughly estimating a good configuration and to provide an intuitive understanding of the relationship between tuning configuration and theta.
A table with the normalized accuracy and latency results as well as an overview of the convex combination of the two can be found in \autoref{apx:accuracyvslatencytradeoff}.
\par
The result of this evaluation of Hyperband is also reflected when comparing all experiments, visualized in \autoref{fig:expoverview}. It shows the mean \ac{HPO} latency (x-axis) and mean error (y-axis) of all configurations that were evaluated, with an ideal configuration being at the origin, with no latency and no error. Since this can be regarded as impossible, we are instead looking for a configuration that approximates this ideal, meaning a configuration that is as close as possible to the bottom left. 
\newline
With the overall lowest (best) mean \ac{avg-wQL}, Hyperband with 30 maximum training jobs can be seen on the bottom center. The overall winner when it comes to latency is the configuration not using hyperparameter tuning (without\_HPO), which can be found at the middle left. Viable Trade-offs between these two extremes are Hyperband with 5 or 20 \texttt{max\_training\_jobs}. 
\newline
The accuracy biased trade-off, namely Hyperband with 30 \texttt{max\_training\_jobs}, is competitive with state of the art commercial \ac{AutoML} forecasting solutions. The custom ensemble pipeline used for experiments in this paper beats Amazon Forecast \cite{noauthor_time_nodate} on the four datasets, with the error (\ac{avg-wQL}) on average being 3.5 \% lower and the end-to-end latency being 16.0 \% lower. 

\section{Conclusion}
The rising demand for \ac{AutoML} in the forecasting space makes the optimisation of general purpose forecasting ensembles important. This paper explores the impact of adding hyperparameter tuning to the deep learning algorithms of such an \ac{AutoML} forecast ensemble, on its performance. It includes an empirical evaluation of the trade-off between different accuracy and latency associated with different tuning strategies and configurations of those. 
\par
A direct comparison between the ensemble performance with and without \ac{HPO} shows that there are effects on both error and latency, with \ac{HPO} lowering the former by 6.1 \% and increasing the latter by 14.0 \% (\autoref{exp:compindensemble}). These changes are the result of an arbitrarily chosen tuning strategy and configuration, merely indicating that hyperparameter tuning can have a positive effect on ensemble accuracy. 
\newline
Further investigations into different hyperparameter tuning strategies show that Hyperband and Bayesian Optimisation have similar pipeline accuracy, with the former having an insignificantly (3.1 \%) lower mean error. Further, the paper shows that Hyperband is on average 32.7 \% faster than Bayesian Optimisation, making it overall favorable for use in the \ac{AutoML} forecast ensemble. 
The analysis and evaluation of different Hyperband configurations suggests that multiple optimal configurations exist, depending on the trade-off between accuracy and latency. With \autoref{fig:thetatradeoff} and \autoref{for:optimalconfig}, the paper provides a comprehensive tool for choosing an optimal configuration for a particular use-case. For the forecasting ensemble setup used in the experiments (\autoref{sec:expensemble}), Hyperband with 5 or 30 trained models for tuning are the best configurations of the forecast ensemble with \ac{HPO}, though a high priority on low latency makes the pipeline setup without \ac{HPO} optimal. 
\newline
The experiments done show that Hyperband with 5 models trained during tuning, leads to a 2.8 \% decrease in error and a 56.6 \% increase in pipeline latency over not using \ac{HPO}, whereas Hyperband with 30 models trained during tuning, leads to a 9.9 \% decrease in error and a 65.8 \% increase in latency over not using \ac{HPO}. As noted in \autoref{apx:enemblerasresults}, results for end-to-end pipeline latency may be sewed by outliers making the comparison, in the experiment section, based on \ac{HPO} latency more reliable.
\par
Statistical significance tests were run for all comparisons made as part of this paper, leading to some of the results being insignificant. This is expected to be primarily due to a low sample size of only 3 repetitions with different random seeds per dataset (12 per configuration tested), making p-values unreliable, even when using non-parametric significance test such as the Wilcoxon Rank Sum test. Instead, the paper uses mean value and standard deviation to compare different experiments. 
\par
The suggested best configuration (with a bias toward low error) resulting form this comparison, namely Hyperband with 30 models trained for tuning, beats the state of the art commercial \ac{AutoML} forecasting solution Amazon forecast \cite{noauthor_time_nodate}, with a 3.5 \% lower error and 16.0 \% lower latency.
\newline
There is still room for improvement, leading to a variety of possible next steps. Firstly, it would make sense to increase the number of repetitions (random seeds) per configuration (e.g. up to 30) to make results more robust and to justify the use of significance tests. Furthermore, it would be recommendable to apply the same experimental design and evaluation methods to a broader range of configurations, possibly also exploring parallelization factor \texttt{max\_parallel\_jobs}. 

% Even though the latter would lead to a more complex evaluation process, it as well as the other suggestions may have the potential to yield additional performance improvements.
\newpage

\appendix
\section{Individual Algorithm Raw Results}
\label{apx:indalgrawresults}

\begin{table}[h]
\centering
\begin{tabular}{|c|c|c|c|c|c|}
\hline
\textbf{Algorithm} & \textbf{Tuning} & \textbf{Dataset} & \textbf{avg\_wQL} & \textbf{Latency} \\
\hline
CNN-QR & hpo & covid\_death & 0.032450 & 2631.378 \\
\hline
CNN-QR & hpo & solar & 0.716921 & 3395.957 \\
\hline
CNN-QR & hpo & electricity & 0.074463 & 3799.176 \\
\hline
CNN-QR & hpo & kaggle\_retail & 0.079736 & 3181.538 \\
\hline
CNN-QR & default & covid\_death & 0.051333 & 1370.223 \\
\hline
CNN-QR & default & solar & 0.757817 & 2685.213 \\
\hline
CNN-QR & default & electricity & 0.073002 & 2354.030 \\
\hline
CNN-QR & default & kaggle\_retail & 0.085578 & 2133.942 \\
\hline
Deep\_AR\_Plus & hpo & covid\_death & 0.057069 & 3310.533 \\
\hline
Deep\_AR\_Plus & hpo & solar & 0.391554 & 12168.680 \\
\hline
Deep\_AR\_Plus & hpo & electricity & 0.055844 & 4427.973 \\
\hline
Deep\_AR\_Plus & hpo & kaggle\_retail & 0.063246 & 3948.480 \\
\hline
Deep\_AR\_Plus & default & covid\_death & 0.108800 & 1939.446 \\
\hline
Deep\_AR\_Plus & default & solar & 0.486347 & 6179.651 \\
\hline
Deep\_AR\_Plus & default & electricity & 0.055448 & 3337.324 \\
\hline
Deep\_AR\_Plus & default & kaggle\_retail & 0.069957 & 2396.073 \\
\hline
\end{tabular}
\caption{Individual Algorithm Raw Results}
\label{tab:indalgrawres}
\end{table}

\section{Individual Algorithm Metric Correlation}
\label{apx:indalgmetriccorr}

To determine a representative metric for the experiments, that were part of the individual algorithm comparison with and without \ac{HPO}, it makes sense to use a metric which is the most representative of different aspects of the forecasting model. Assuming that these aspects are reflected in the metrics computed for each experiment, namely \ac{MAPE}, \ac{MASE}, \ac{WAPE}, \ac{wQL-90}, \ac{wQL-50}, \ac{wQL-10} and \ac{avg-wQL}, this means it is coherent to find a metric which correlates highly with all other metrics. To achieve this I constructed a data frame with the respective metrics resulting form all 16 experiment part of \autoref{sec:expindalgs}. In the following I computed the standard pearson correlation coefficient, given by \autoref{for:pearson}, between the observations of these different metrics. 

\begin{equation}
    \rho_{X,Y} = \frac{cov(X,Y)}{\sigma X \sigma Y}
    \label{for:pearson}
\end{equation}

This measures linear correlation between pairs of metrics. When done for all pairs of metrics this results in a two dimensional matrix as given by \autoref{tab:metriccorr}. Note that the table also nicely shows that \ac{wQL-50} and median based \ac{WAPE} are identical. The mean for each column in this table, can be interpreted as the representativity of the respective metric. This measure it highest (best) for the \ac{wQL-90} with a mean correlation of 0.828 and second highest for \ac{avg-wQL} with a mean correlation of 0.823. The mean correlation for all other metrics can be derived form the bottom row of \autoref{tab:metriccorr}.

\begin{table}[h]
    \centering
    \begin{tabular}{|c|ccccccc|}
        \hline
        & \ac{MAPE} & \ac{MASE} & \ac{WAPE} & \ac{wQL-90} & \ac{wQL-50} & \ac{wQL-10} & \ac{avg-wQL} \\
        \hline
        \ac{MAPE} & 1.000 & -0.223 & 0.992 & 0.978 & 0.993 & 0.950 & 0.981 \\
        \ac{MASE} & -0.223 & 1.000 & -0.200 & -0.123 & -0.199 & -0.200 & -0.192 \\
        \ac{WAPE} & 0.992 & -0.200 & 1.000 & 0.994 & 1.000 & 0.971 & 0.995 \\
        \ac{wQL-90} & 0.978 & -0.123 & 0.994 & 1.000 & 0.994 & 0.964 & 0.990 \\
        \ac{wQL-50} & 0.993 & -0.199 & 1.000 & 0.994 & 1.000 & 0.968 & 0.994 \\
        \ac{wQL-10} & 0.950 & -0.200 & 0.971 & 0.964 & 0.968 & 1.000 & 0.989 \\
        \ac{avg-wQL} & 0.981 & -0.192 & 0.995 & 0.990 & 0.994 & 0.989 & 1.000 \\
        \hline
        mean-$\rho$ & 0.810 & -0.020 & 0.822 & 0.828 & 0.821 & 0.806 & 0.823 \\
        \hline
    \end{tabular}
    \caption{Individual Algorithm Metric Correlation}
    \label{tab:metriccorr}
\end{table}

The choice of \ac{avg-wQL} as the comparison metric results form its high mean correlation and the fact that is approximates a strictly proper scoring rule as outlined in  \autoref{sec:forecastingalgs}, making it both representative of the other metrics and ensuring that the forecaster is making a careful and honest prediction.

\section{Forecast Ensemble Raw Results}
\label{apx:enemblerasresults}

This section contains the raw results of all experiments conducted as part of this paper. The Table contains the name of the experiment, made up of the strategy name and the number of models trained during tuning (if \ac{HPO} was used). Each such configuration was evaluated with three different random seeds indicated by the Version field for the four different datasets. The error column refers to the \ac{avg-wQL} of the experiment and the Pipeline and \ac{HPO} columns refer to the respective latency. It is important that values for Pipeline latency may contain outliers that were caused by resource allocation limits. These were only discovered in later analysis, which is the reason why \ac{HPO} Latency was used for most of the comparisons.  
\begin{table}[H]
\centering
\begin{tabular}{|c|c|c|c|c|c|c|}
    \hline
    \textbf{Experiment} & \textbf{Dataset} & \textbf{Version} & \textbf{Error} & \textbf{Pipeline} & \textbf{\ac{HPO}}\\
    \hline
Bayesian\_iter\_10 & covid\_death & a & 0.0263 & 4223.7 & 1780.6 \\
Bayesian\_iter\_10 & covid\_death & b & 0.0269 & 3927.1 & 1237.6 \\
Bayesian\_iter\_10 & covid\_death & c & 0.0277 & 5661.5 & 1280. \\
Bayesian\_iter\_10 & electricity & a & 0.0581 & 6344.7 & 3339.5 \\
Bayesian\_iter\_10 & electricity & b & 0.0551 & 6183.9 & 3427.7 \\
Bayesian\_iter\_10 & electricity & c & 0.0588 & 8166.3 & 4548.7 \\
Bayesian\_iter\_10 & kaggle\_retail & a & 0.078 & 3767.4 & 1474.7 \\
Bayesian\_iter\_10 & kaggle\_retail & b & 0.0806 & 3561.2 & 1178.5 \\
Bayesian\_iter\_10 & kaggle\_retail & c & 0.0789 & 5645.6 & 1280.0 \\
Bayesian\_iter\_10 & solar & a & 0.5004 & 9772.5 & 4624.0 \\
Bayesian\_iter\_10 & solar & b & 0.5425 & 10263.8 & 5429.7 \\
Bayesian\_iter\_10 & solar & c & 0.5419 & 9640.4 & 4580.4 \\
Bayesian\_iter\_15 & covid\_death & a & 0.0269 & 4624.4 & 1973.3 \\
Bayesian\_iter\_15 & covid\_death & b & 0.027 & 5541.7 & 2855.4 \\
Bayesian\_iter\_15 & covid\_death & c & 0.0266 & 4317.0 & 1830. \\
Bayesian\_iter\_15 & electricity & a & 0.0586 & 8697.3 & 5322.2 \\
Bayesian\_iter\_15 & electricity & b & 0.0589 & 33789.0 & 4640.4 \\
Bayesian\_iter\_15 & electricity & c & 0.0598 & 8103.5 & 4982.7 \\
Bayesian\_iter\_15 & kaggle\_retail & a & 0.0807 & 4393.6 & 2130.7 \\
Bayesian\_iter\_15 & kaggle\_retail & b & 0.0679 & 4032.1 & 1384.7 \\
Bayesian\_iter\_15 & kaggle\_retail & c & 0.0767 & 4746.6 & 2376. \\
Bayesian\_iter\_15 & solar & a & 0.4125 & 10204.3 & 5767.9 \\
Bayesian\_iter\_15 & solar & b & 0.4968 & 33781.5 & 5100.8 \\
Bayesian\_iter\_15 & solar & c & 0.4976 & 12231.3 & 7029.0 \\
Bayesian\_iter\_20 & covid\_death & a & 0.027 & 4343.5 & 1979.9 \\
Bayesian\_iter\_20 & covid\_death & b & 0.027 & 4825.9 & 1967.3 \\
Bayesian\_iter\_20 & covid\_death & c & 0.0269 & 7010.5 & 4343.1 \\
Bayesian\_iter\_20 & electricity & a & 0.0593 & 9853.5 & 6420.2 \\
Bayesian\_iter\_20 & electricity & b & 0.0592 & 8178.6 & 5257.1 \\
Bayesian\_iter\_20 & electricity & c & 0.058 & 8682.1 & 5845.7 \\
Bayesian\_iter\_20 & kaggle\_retail & a & 0.0736 & 4354.5 & 2018.6 \\
Bayesian\_iter\_20 & kaggle\_retail & b & 0.0769 & 4648.1 & 2082.4 \\
Bayesian\_iter\_20 & kaggle\_retail & c & 0.071 & 6467.9 & 4248.1 \\
Bayesian\_iter\_20 & solar & a & 0.5514 & 17979.9 & 9654.2 \\
Bayesian\_iter\_20 & solar & b & 0.4412 & 14776.0 & 7107.7 \\
Bayesian\_iter\_20 & solar & c & 0.5262 & 16592.3 & 10359.2 \\
    \hline
\end{tabular}
\caption{Raw Ensemble Results 1}
\label{tab:rawensemble1}
\end{table}

\begin{table}[H]
\centering
\begin{tabular}{|c|c|c|c|c|c|c|}
    \hline
    \textbf{Experiment} & \textbf{Dataset} & \textbf{Version} & \textbf{Error} & \textbf{Pipeline} & \textbf{\ac{HPO}}\\
    \hline
Hyperband\_iter\_10 & covid\_death & a & 0.0267 & 3587.4 & 1140. \\
Hyperband\_iter\_10 & covid\_death & b & 0.0275 & 3798.3 & 1420.1 \\
Hyperband\_iter\_10 & covid\_death & c & 0.0257 & 11357.2 & 2082.7 \\
Hyperband\_iter\_10 & electricity & a & 0.0611 & 4258.5 & 1293. \\
Hyperband\_iter\_10 & electricity & b & 0.059 & 5704.7 & 2751.3 \\
Hyperband\_iter\_10 & electricity & c & 0.0597 & 12019.9 & 2623.4 \\
Hyperband\_iter\_10 & kaggle\_retail & a & 0.0725 & 3246.2 & 1182.4 \\
Hyperband\_iter\_10 & kaggle\_retail & b & 0.0816 & 3540.2 & 1382.7 \\
Hyperband\_iter\_10 & kaggle\_retail & c & 0.0673 & 10332.5 & 2034.3 \\
Hyperband\_iter\_10 & solar & a & 0.5467 & 9983.5 & 3516.9 \\
Hyperband\_iter\_10 & solar & b & 0.4766 & 7990.5 & 3731.4 \\
Hyperband\_iter\_10 & solar & c & 0.4596 & 18505.6 & 5805.6 \\
Hyperband\_iter\_15 & covid\_death & a & 0.027 & 6341.1 & 3796.7 \\
Hyperband\_iter\_15 & covid\_death & b & 0.027 & 4346.5 & 1925.0 \\
Hyperband\_iter\_15 & covid\_death & c & 0.027 & 4205.1 & 1454.4 \\
Hyperband\_iter\_15 & electricity & a & 0.0601 & 7147.0 & 3748. \\
Hyperband\_iter\_15 & electricity & b & 0.0602 & 6197.5 & 3194.6 \\
Hyperband\_iter\_15 & electricity & c & 0.0596 & 6087.5 & 2223.2 \\
Hyperband\_iter\_15 & kaggle\_retail & a & 0.0805 & 6133.3 & 3754.1 \\
Hyperband\_iter\_15 & kaggle\_retail & b & 0.0696 & 4443.8 & 2183.6 \\
Hyperband\_iter\_15 & kaggle\_retail & c & 0.0596 & 3648.4 & 1311.3 \\
Hyperband\_iter\_15 & solar & a & 0.4914 & 7422.6 & 3776. \\
Hyperband\_iter\_15 & solar & b & 0.4506 & 8389.0 & 3391.0 \\
Hyperband\_iter\_15 & solar & c & 0.5778 & 7546.2 & 2945.7 \\
Hyperband\_iter\_20 & covid\_death & a & 0.0267 & 3560.1 & 1320.4 \\
Hyperband\_iter\_20 & covid\_death & b & 0.0262 & 4117.6 & 1601.5 \\
Hyperband\_iter\_20 & covid\_death & c & 0.0269 & 4222.5 & 1698.8 \\
Hyperband\_iter\_20 & electricity & a & 0.0605 & 6008.4 & 2774.5 \\
Hyperband\_iter\_20 & electricity & b & 0.0587 & 6475.5 & 2841.5 \\
Hyperband\_iter\_20 & electricity & c & 0.0606 & 8050.1 & 5114.0 \\
Hyperband\_iter\_20 & kaggle\_retail & a & 0.0665 & 4553.6 & 2178.5 \\
Hyperband\_iter\_20 & kaggle\_retail & b & 0.07 & 3337.0 & 1240.4 \\
Hyperband\_iter\_20 & kaggle\_retail & c & 0.0783 & 4494.1 & 2271.0 \\
Hyperband\_iter\_20 & solar & a & 0.4208 & 9092.8 & 3919.0 \\
Hyperband\_iter\_20 & solar & b & 0.5746 & 7143.6 & 2902.2 \\
Hyperband\_iter\_20 & solar & c & 0.4481 & 8658.0 & 3843.2 \\
Hyperband\_iter\_30 & covid\_death & a & 0.0271 & 4438.9 & 1725.3 \\
Hyperband\_iter\_30 & covid\_death & b & 0.027 & 4435.8 & 1578.7 \\
Hyperband\_iter\_30 & covid\_death & c & 0.0264 & 4592.4 & 1874.2 \\
Hyperband\_iter\_30 & electricity & a & 0.0601 & 5846.7 & 2486.6 \\
Hyperband\_iter\_30 & electricity & b & 0.0568 & 6756.8 & 2951.3 \\
Hyperband\_iter\_30 & electricity & c & 0.0534 & 5846.8 & 2831.2 \\
Hyperband\_iter\_30 & kaggle\_retail & a & 0.0686 & 5500.8 & 3045. \\
Hyperband\_iter\_30 & kaggle\_retail & b & 0.0804 & 3564.8 & 1494.3 \\
Hyperband\_iter\_30 & kaggle\_retail & c & 0.074 & 4451.8 & 2328.0 \\
Hyperband\_iter\_30 & solar & a & 0.3527 & 9879.0 & 4627.8 \\
Hyperband\_iter\_30 & solar & b & 0.4613 & 8103.9 & 2982.2 \\
Hyperband\_iter\_30 & solar & c & 0.537 & 7696.9 & 4023. \\
    \hline
\end{tabular}
\caption{Raw Ensemble Results 2}
\label{tab:rawensemble2}
\end{table}

\begin{table}[H]
\centering
\begin{tabular}{|c|c|c|c|c|c|c|}
    \hline
    \textbf{Experiment} & \textbf{Dataset} & \textbf{Version} & \textbf{Error} & \textbf{Pipeline} & \textbf{\ac{HPO}}\\
    \hline
Hyperband\_iter\_40 & covid\_death & a & 0.0269 & 48021.3 & 4363.9 \\
Hyperband\_iter\_40 & covid\_death & b & 0.0271 & 45606.7 & 3552.7 \\
Hyperband\_iter\_40 & covid\_death & c & 0.0266 & 46251.1 & 3927.3 \\
Hyperband\_iter\_40 & electricity & a & 0.0587 & 48263.1 & 4545.0 \\
Hyperband\_iter\_40 & electricity & b & 0.06 & 49197.5 & 6076. \\
Hyperband\_iter\_40 & electricity & c & 0.0594 & 47365.9 & 4641.6 \\
Hyperband\_iter\_40 & kaggle\_retail & a & 0.0753 & 46228.8 & 3639.4 \\
Hyperband\_iter\_40 & kaggle\_retail & b & 0.0777 & 41334.7 & 3372.4 \\
Hyperband\_iter\_40 & kaggle\_retail & c & 0.0788 & 41088.4 & 3290.2 \\
Hyperband\_iter\_40 & solar & a & 0.6258 & 48382.6 & 5897.5 \\
Hyperband\_iter\_40 & solar & b & 0.6215 & 45575.8 & 3031. \\
Hyperband\_iter\_40 & solar & c & 0.5309 & 51554.3 & 10807.2 \\
Hyperband\_iter\_50 & covid\_death & a & 0.0266 & 6004.1 & 2824.0 \\
Hyperband\_iter\_50 & covid\_death & b & 0.027 & 4712.3 & 2091.5 \\
Hyperband\_iter\_50 & covid\_death & c & 0.0271 & 9410.8 & 3973.9 \\
Hyperband\_iter\_50 & electricity & a & 0.0598 & 6351.7 & 2749.5 \\
Hyperband\_iter\_50 & electricity & b & 0.0548 & 6533.3 & 2438.3 \\
Hyperband\_iter\_50 & electricity & c & 0.0594 & 10278.8 & 5287. \\
Hyperband\_iter\_50 & kaggle\_retail & a & 0.0733 & 5308.2 & 2757.6 \\
Hyperband\_iter\_50 & kaggle\_retail & b & 0.0758 & 4187.5 & 1815.2 \\
Hyperband\_iter\_50 & kaggle\_retail & c & 0.0808 & 6925.2 & 4538.6 \\
Hyperband\_iter\_50 & solar & a & 0.5677 & 8146.5 & 3263.7 \\
Hyperband\_iter\_50 & solar & b & 0.5725 & 33422.1 & 5781.8 \\
Hyperband\_iter\_50 & solar & c & 0.5055 & 11485.9 & 6683. \\
Hyperband\_iter\_5 & covid\_death & a & 0.027 & 5893.6 & 1387.5 \\
Hyperband\_iter\_5 & covid\_death & b & 0.0296 & 5731.7 & 1313.4 \\
Hyperband\_iter\_5 & covid\_death & c & 0.0267 & 3502.4 & 1110.6 \\
Hyperband\_iter\_5 & electricity & a & 0.0583 & 6933.8 & 1980.9 \\
Hyperband\_iter\_5 & electricity & b & 0.0581 & 7050.5 & 3923.0 \\
Hyperband\_iter\_5 & electricity & c & 0.0602 & 5363.3 & 1797.7 \\
Hyperband\_iter\_5 & kaggle\_retail & a & 0.0735 & 5764.4 & 1724.4 \\
Hyperband\_iter\_5 & kaggle\_retail & b & 0.08 & 5629.7 & 1605.7 \\
Hyperband\_iter\_5 & kaggle\_retail & c & 0.0769 & 3546.9 & 1066.8 \\
Hyperband\_iter\_5 & solar & a & 0.4212 & 8402.2 & 4062.4 \\
Hyperband\_iter\_5 & solar & b & 0.5323 & 8090.0 & 3872.7 \\
Hyperband\_iter\_5 & solar & c & 0.5132 & 9347.4 & 3041. \\
without\_HPO & covid\_death & a & 0.0271 & 2640.4 & 0. \\
without\_HPO & covid\_death & b & 0.0266 & 2957.5 & 0. \\
without\_HPO & covid\_death & c & 0.0261 & 2534.0 & 0. \\
without\_HPO & electricity & a & 0.0608 & 3997.0 & 0. \\
without\_HPO & electricity & b & 0.0599 & 3731.3 & 0. \\
without\_HPO & electricity & c & 0.0551 & 3674.3 & 0. \\
without\_HPO & kaggle\_retail & a & 0.0775 & 3376.9 & 0. \\
without\_HPO & kaggle\_retail & b & 0.0737 & 2521.1 & 0. \\
without\_HPO & kaggle\_retail & c & 0.0773 & 2284.0 & 0. \\
without\_HPO & solar & a & 0.4803 & 5744.8 & 0. \\
without\_HPO & solar & b & 0.5289 & 5193.8 & 0. \\
without\_HPO & solar & c & 0.52 & 5719.0 & 0. \\
    \hline
\end{tabular}
\caption{Raw Ensemble Results 3}
\label{tab:rawensemble3}
\end{table}

\newpage
\section{Accuracy vs. Latency Trade-Off}
\label{apx:accuracyvslatencytradeoff}

To evaluate different trade-offs between accuracy (or error) and latency (or cost), both are normalized, meaning that the sum of all observations per variable is equal to 1. This is done by dividing each value by the sum of all values, for all Hyperband configuration tests. The result is given by \autoref{tab:latencyanderrornorm}.

\begin{table}[ht]
    \centering 
    \begin{tabular}{|c|c|c|}
    \hline
    \textbf{Configuration} & \textbf{Error} & \textbf{Latency} \\
    \hline
        5 & 0.139120 & 0.104813 \\
        10 & 0.139633 & 0.112077 \\
        15 & 0.141495 & 0.128731 \\
        20 & 0.136359 & 0.118955 \\
        30 & 0.130653 & 0.143418 \\
        40 & 0.161295 & 0.220888 \\
        50 & 0.151445 & 0.171117 \\
    \hline
        sum & 1 & 1 \\
    \hline
    \end{tabular}
    \caption{Normalized Error and Latency}
    \label{tab:latencyanderrornorm}
\end{table}

The following \autoref{tab:accuracyvslatencytradeoff} shows the cost associated with different convex combinations of normalized error and latency from \autoref{tab:latencyanderrornorm} for Hyperband with the maximum number of training jobs \texttt{max\_training\_jobs} $\in \{5,10,15,20,30,40,50\}$ for different $\theta$. The configuration with the lowest cost per row is bold, to increase readability. In the continuous case, if $\theta_0 \le 0.1798$ the configuration Hyperband with \texttt{max\_training\_jobs} 30 is optimal and for $\theta_0 > 0.1798$ the configuration Hyperband with \texttt{max\_training\_jobs} 5 is optimal.

\begin{table}[ht]
    \centering
    \caption{Cost of Convex Accuracy \& Latency Combination}
    \begin{tabular}{|c|ccccccc|}
        \hline
         & \multicolumn{7}{c|}{Number of Trained Models per Tuning Job} \\
        \hline
        $\theta$ & \textbf{5} & \textbf{10} &  \textbf{15} & \textbf{20} & \textbf{30} & \textbf{40} & \textbf{50} \\
        \hline
        0.0 & 0.13912 & 0.13963 & 0.14149 & 0.13636 & \textbf{0.13065} & 0.1613 & 0.15145 \\
        0.04 & 0.13775 & 0.13853 & 0.14098 & 0.13566 & \textbf{0.13116} & 0.16368 & 0.15223 \\
        0.08 & 0.13638 & 0.13743 & 0.14047 & 0.13497 & \textbf{0.13167} & 0.16606 & 0.15302 \\
        0.12 & 0.135 & 0.13633 & 0.13996 & 0.13427 & \textbf{0.13218} & 0.16845 & 0.15381 \\
        0.16 & 0.13363 & 0.13522 & 0.13945 & 0.13357 & \textbf{0.1327} & 0.17083 & 0.15459 \\
        0.2 & \textbf{0.13226} & 0.13412 & 0.13894 & 0.13288 & 0.13321 & 0.17321 & 0.15538 \\
        0.24 & \textbf{0.13089} & 0.13302 & 0.13843 & 0.13218 & 0.13372 & 0.1756 & 0.15617 \\
        0.28 & \textbf{0.12951} & 0.13192 & 0.13792 & 0.13149 & 0.13423 & 0.17798 & 0.15695 \\
        0.32 & \textbf{0.12814} & 0.13081 & 0.13741 & 0.13079 & 0.13474 & 0.18037 & 0.15774 \\
        0.36 & \textbf{0.12677} & 0.12971 & 0.1369 & 0.13009 & 0.13525 & 0.18275 & 0.15853 \\
        0.4 & \textbf{0.1254} & 0.12861 & 0.13639 & 0.1294 & 0.13576 & 0.18513 & 0.15931 \\
        0.44 & \textbf{0.12402} & 0.12751 & 0.13588 & 0.1287 & 0.13627 & 0.18752 & 0.1601 \\
        0.48 & \textbf{0.12265} & 0.12641 & 0.13537 & 0.12801 & 0.13678 & 0.1899 & 0.16089 \\
        0.52 & \textbf{0.12128} & 0.1253 & 0.13486 & 0.12731 & 0.13729 & 0.19228 & 0.16167 \\
        0.56 & \textbf{0.11991} & 0.1242 & 0.13435 & 0.12661 & 0.1378 & 0.19467 & 0.16246 \\
        0.6 & \textbf{0.11854} & 0.1231 & 0.13384 & 0.12592 & 0.13831 & 0.19705 & 0.16325 \\
        0.64 & \textbf{0.11716} & 0.122 & 0.13333 & 0.12522 & 0.13882 & 0.19943 & 0.16404 \\
        0.68 & \textbf{0.11579} & 0.12089 & 0.13282 & 0.12452 & 0.13933 & 0.20182 & 0.16482 \\
        0.72 & \textbf{0.11442} & 0.11979 & 0.1323 & 0.12383 & 0.13984 & 0.2042 & 0.16561 \\
        \hline
    \end{tabular}
    \label{tab:accuracyvslatencytradeoff}
\end{table}

\begin{table}[H]
    \centering
    \caption{Cost of Convex Accuracy \& Latency Combination}
    \begin{tabular}{|c|ccccccc|}
        \hline
         & \multicolumn{7}{c|}{Number of Trained Models per Tuning Job} \\
        \hline
        $\theta$ & \textbf{5} & \textbf{10} &  \textbf{15} & \textbf{20} & \textbf{30} & \textbf{40} & \textbf{50} \\
        \hline
        0.76 & \textbf{0.11305} & 0.11869 & 0.13179 & 0.12313 & 0.14035 & 0.20659 & 0.1664 \\
        0.8 & \textbf{0.11167} & 0.11759 & 0.13128 & 0.12244 & 0.14087 & 0.20897 & 0.16718 \\
        0.84 & \textbf{0.1103} & 0.11649 & 0.13077 & 0.12174 & 0.14138 & 0.21135 & 0.16797 \\
        0.88 & \textbf{0.10893} & 0.11538 & 0.13026 & 0.12104 & 0.14189 & 0.21374 & 0.16876 \\
        0.92 & \textbf{0.10756} & 0.11428 & 0.12975 & 0.12035 & 0.1424 & 0.21612 & 0.16954 \\
        0.96 & \textbf{0.10619} & 0.11318 & 0.12924 & 0.11965 & 0.14291 & 0.2185 & 0.17033 \\
        1.0 & \textbf{0.10481} & 0.11208 & 0.12873 & 0.11896 & 0.14342 & 0.22089 & 0.17112 \\
        1.04 & \textbf{0.10344} & 0.11098 & 0.12822 & 0.11826 & 0.14393 & 0.22327 & 0.1719 \\
        1.08 & \textbf{0.10207} & 0.10987 & 0.12771 & 0.11756 & 0.14444 & 0.22566 & 0.17269 \\
        1.12 & \textbf{0.1007} & 0.10877 & 0.1272 & 0.11687 & 0.14495 & 0.22804 & 0.17348 \\
        1.16 & \textbf{0.09932} & 0.10767 & 0.12669 & 0.11617 & 0.14546 & 0.23042 & 0.17426 \\
        1.2 & \textbf{0.09795} & 0.10657 & 0.12618 & 0.11547 & 0.14597 & 0.23281 & 0.17505 \\
        1.24 & \textbf{0.09658} & 0.10546 & 0.12567 & 0.11478 & 0.14648 & 0.23519 & 0.17584 \\
        1.28 & \textbf{0.09521} & 0.10436 & 0.12516 & 0.11408 & 0.14699 & 0.23757 & 0.17662 \\
        1.32 & \textbf{0.09384} & 0.10326 & 0.12465 & 0.11339 & 0.1475 & 0.23996 & 0.17741 \\
        1.36 & \textbf{0.09246} & 0.10216 & 0.12414 & 0.11269 & 0.14801 & 0.24234 & 0.1782 \\
        1.4 & \textbf{0.09109} & 0.10106 & 0.12363 & 0.11199 & 0.14852 & 0.24473 & 0.17899 \\
        1.44 & \textbf{0.08972} & 0.09995 & 0.12311 & 0.1113 & 0.14903 & 0.24711 & 0.17977 \\
        1.48 & \textbf{0.08835} & 0.09885 & 0.1226 & 0.1106 & 0.14955 & 0.24949 & 0.18056 \\
        1.52 & \textbf{0.08697} & 0.09775 & 0.12209 & 0.10991 & 0.15006 & 0.25188 & 0.18135 \\
        1.56 & \textbf{0.0856} & 0.09665 & 0.12158 & 0.10921 & 0.15057 & 0.25426 & 0.18213 \\
        \hline
    \end{tabular}
    \label{tab:accuracyvslatencytradeoff2}
\end{table}
\null
\vfill

%Bibliography
\bibliographystyle{unsrt}  
\bibliography{references}

\end{document}